%% file: main.tex
\documentclass[10pt,journal,compsoc]{IEEEtran}

\usepackage[nocompress]{cite}

\usepackage[accsupp]{axessibility}  %

\usepackage{graphicx}
\usepackage{amsmath}
\usepackage{amssymb}
\usepackage{booktabs}

\usepackage[pagebackref,breaklinks,colorlinks]{hyperref}

\usepackage{amsmath,amsfonts}
\usepackage{algorithmic}
\usepackage{algorithm}
\usepackage{array}
\usepackage[caption=false,font=normalsize,labelfont=sf,textfont=sf]{subfig}
\usepackage{textcomp}
\usepackage{stfloats}
\usepackage{url}
\usepackage{verbatim}
\usepackage{graphicx}
\usepackage{bbding}
\usepackage{cite}
\usepackage{caption}
\usepackage{float}
\usepackage{subfig}

\usepackage{booktabs}
\usepackage{multirow}
\usepackage{graphicx}
\usepackage{array}
\usepackage{xcolor}
\usepackage{booktabs}
\usepackage{multirow}
\usepackage{wrapfig} 
\usepackage{booktabs}
\usepackage{xspace}

\usepackage{arydshln}

\newcommand{\ie}{\textit{i.e.,}\xspace}
\newcommand{\eg}{\textit{e.g.,}\xspace}

\newcommand{\yesbut}{{\textsc{YesBut}}\xspace}

\begin{document}

\title{When `\textsc{Yes}' Meets `\textsc{But}': Can Large Models Comprehend Contradictory Humor Through Comparative Reasoning?}

\author{Tuo~Liang$^{1*}$,
        Zhe~Hu$^{2*}$,
        Jing Li$^2$, 
        Hao Zhang$^1$,
        Yiren Lu$^1$, 
        Yunlai Zhou$^1$, 
        Yiran Qiao$^1$, \\\vspace{-2mm}
        Disheng Liu$^1$,
        Jeirui Peng$^1$,
        Jing Ma$^1$,
        Yu~Yin$^1$\textsuperscript{\Envelope} \\\vspace{2mm}
        \url{https://vulab-ai.com/projects/yesbut-v2/}
\IEEEcompsocitemizethanks{\IEEEcompsocthanksitem Tuo~Liang, Hao Zhang, Yiren Lu, Yunlai Zhou, Yiran Qiao, Disheng Liu, Jeirui Peng, Jing Ma, Yu~Yin are with Computer and Data Sciences Department, Case Western Reserve University. Email: tuo.liang@case.edu
\IEEEcompsocthanksitem Zhe~Hu and Jing Li are with Department of Computing, The Hong Kong Polytechnic University. Email: zhe-derek.hu@connect.polyu.hk 
}
\thanks{* These authors contribute equally to this research.}
\thanks{\textsuperscript{\Envelope} Correspondence to: Yu Yin (yu.yin@case.edu).}}

\IEEEtitleabstractindextext{
\input{abstract}
}

\maketitle



\ifCLASSOPTIONcompsoc

\input{introduction}

\input{related_work}

\input{dataset}
\input{experiment}
\input{results}

\input{Analysis}

\input{conclusion}

\bibliographystyle{IEEEtran}
\bibliography{refer}

\ifCLASSOPTIONcaptionsoff
  \newpage
\fi

\newpage
\input{appendix}

\vfill

\end{document}

%% file: abstract.tex
\begin{abstract}
Understanding humor—particularly when it involves complex, contradictory narratives that require comparative reasoning—remains a significant challenge for large vision-language models (VLMs). This limitation hinders AI’s ability to engage in human-like reasoning and cultural expression.
In this paper, we investigate this challenge through an in-depth analysis of comics that juxtapose panels to create humor through contradictions. We introduce the \yesbut (V2), a novel benchmark with 1,262 comic images from diverse multilingual and multicultural contexts, featuring comprehensive annotations that capture various aspects of narrative understanding. 
Using this benchmark, we systematically evaluate a wide range of VLMs through four complementary tasks spanning from surface content comprehension to deep narrative reasoning, with particular emphasis on comparative reasoning between contradictory elements.
Our extensive experiments reveal that even the most advanced models significantly underperform compared to humans, with common failures in visual perception, key element identification, comparative analysis and hallucinations. We further investigate text-based training strategies and social knowledge augmentation methods to enhance model performance. Our findings not only highlight critical weaknesses in VLMs' understanding of cultural and creative expressions but also provide pathways toward developing context-aware models capable of deeper narrative understanding though comparative reasoning.

\end{abstract}

%% file: introduction.tex
\begin{figure}[t]
    \centering
    \includegraphics[width=1\linewidth]{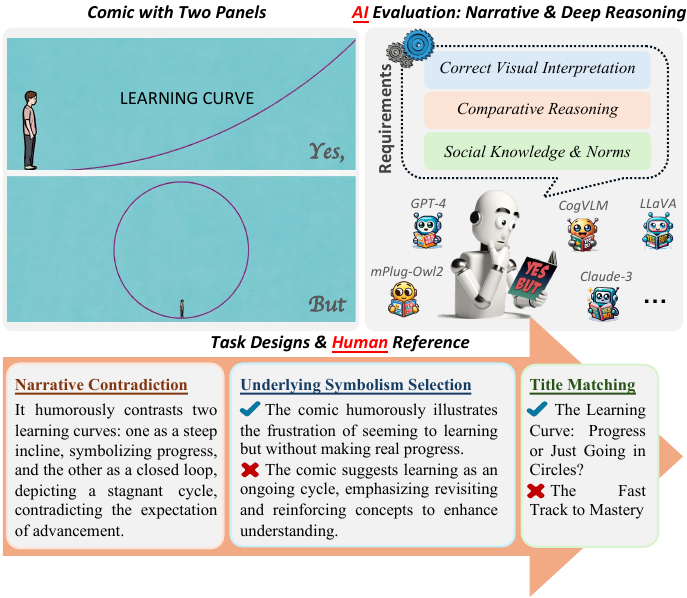}
    \caption{
    We introduce the \yesbut (V2), a benchmark for assessing AI's ability to interpret juxtaposed comic panels with contradictory narratives. Unlike existing benchmarks, it emphasizes visual understanding, comparative reasoning, and social knowledge. To capture the layered reasoning required for interpreting these contradictions, we design multi-tiered tasks—ranging from basic content recognition to deep narrative comprehension—ensuring a comprehensive assessment of AI’s interpretative abilities.
    }
    \label{fig:intro_example}
\end{figure}
\section{Introduction}


\IEEEPARstart{C}{omics} are more than just entertainment—they are intricate puzzles of visual narratives that merge images, text, and spatial relationships to convey layered meanings. Unlike novels, which rely solely on text, or videos, which typically follow a linear progression, comics often use fragmented yet interconnected panels to construct meaning through juxtaposition, requiring sophisticated comparative reasoning to interpret.
While humans naturally interpret this nonlinear structure due to their cognitive abilities for comparison and cultural knowledge, current vision-language models (VLMs) struggle to replicate such nuanced understanding and comparative reasoning~\cite{hu2023language,hessel-etal-2023-androids,rayhan2023artificial}.

Evaluating how effectively VLMs perform comparative reasoning to interpret visual juxtaposition is essential for developing socially intelligent AI systems~\cite{koivisto2023best}
Comics encapsulate complex human emotions, cultural nuances, and symbolic storytelling~\cite{duncan2009power}, making them a rigorous testbed for assessing AI’s interpretive and comparative capabilities. A deeper understanding of visual juxtaposition can enhance scene interpretation, facilitate AI-driven storytelling that resonates with human audiences, and improve creative content generation in a more human-like manner.
Addressing the challenges of visual juxtaposition allows VLMs to move beyond basic pattern recognition toward nuanced, context-aware reasoning, fostering deeper and more meaningful human–AI interactions.

While previous studies~\cite{hessel-etal-2023-androids,yang2024can} have explored humor understanding with VLMs, most studies have focused on single-panel comics, where the narrative is self-contained within a single image. In contrast, our work examines how VLMs handle juxtaposition—a technique that places contrasting elements side by side to create humor, irony, or thought-provoking narratives~\cite{young2003art,groensteen2013comics}. This approach challenges VLMs to engage in \textit{comparative reasoning and analysis} to decipher the intricate relationships between panels to capture the overall narrative~\cite{bearne2003rethinking,dittmer2010comic,schechter2011juxtaposition}.

In this work, we explore whether AI can recognize and interpret implicit contradictions in juxtaposed panels that contribute to meaning and humor. To illustrate, consider the comic in Fig.~\ref{fig:intro_example}, which humorously illustrates the concept of a "learning curve." In the "Yes" panel, a man observes a steady upward curve labeled "LEARNING CURVE," symbolizing smooth progress. However, in the "But" panel, the curve twists into a closed loop, trapping him inside, humorously depicting the illusion of learning without real progress. This example highlights the need for deep contextual understanding and comparative reasoning—capabilities that remain limited in current models due to the constraints of the autoregressive paradigm, which hinders bidirectional reasoning~\cite{kuttner2021comics,tong2023eliminating,bubeck2023sparks}.

We identify three primary challenges in achieving robust juxtaposition understanding in comics (Figure~\ref{fig:intro_example}). The first challenge is \textbf{accurate visual interpretation}, which involves decoding the visual elements within each panel. The second is \textbf{comparative reasoning}, which requires integrating and comparing key elements across multiple panels to detect contradictions that create humor or irony in the overall narrative.
Finally, \textbf{social and cultural comprehension} is essential, as recognizing subtle social cues, conventions, and cultural contexts significantly influences the interpretation of comic humor and emotional responses.
To this end, we introduce \yesbut, the first benchmark designed to assess VLMs' capacity for recognizing humor through juxtaposition and contradiction. Our dataset is uniquely annotated to capture multiple layers of narrative complexity. Each sample includes a literal description of the scene, an explicit contradiction statement that clarifies the humorous contrast, the underlying symbolism or message conveyed by the comic, a title summarizing the overall theme, and relevant background knowledge—including social norms, cultural references, and linguistic context—essential for full interpretation of the underlying humor.


Based on our annotations, we propose four complementary tasks  that progressively assess comic understanding across different cognitive levels: (1)
Literal Description Writing evaluates a model’s perceptual ability by requiring to generate a surface-level description with an explicit depiction of the visual and textual content presented in the comic, 
(2) Contradiction Generation focuses on identifying and articulating the core contradiction in the narrative that often serves as the basis for humor; 
(3) Underlying Symbolism Selection measures deeper interpretative reasoning by challenging models to infer the abstract message or commentary embedded in the comic;
and (4) Title Matching assesses holistic understanding and thematic summarization by requiring models to select a title that accurately encapsulates the comic's thematic essence.
This hierarchical evaluation  establishes a systematic approach for measuring progress in machine humor comprehension while highlighting specific reasoning capabilities needed for more sophisticated semantic understanding of visual narratives.


Building on our preliminary work~\cite{hu2024yesbut}, which introduced a small-scale dataset of 349 images and identified significant gaps in VLMs' comic narrative understanding, this paper presents a substantial expansion and refinement. We evaluate a wide range of models including both recent LLMs and VLMs using our extended benchmark. The experiments as well as extensive analysis reveal that current large models still face challenges for contradictory humor understanding with comparative reasoning, and tend to make common errors including visual perception, key element identification, and hallucination in narrative understanding. Our analysis further show that 
augmenting social knowledge and model training through textual data distillation can improve model performance. Our key contributions are:
\begin{itemize}
  \item \textbf{Larger and More Diverse Dataset}: We expand \textsc{YesBut} from 349 to 1,262 images, improving the benchmark’s robustness and diversity to facilitate a more comprehensive evaluation of VLM capabilities.
  The comics encompass multi-cultural backgrounds, enabling richer assessment of linguistic and cultural influences on humor comprehension.
  \item \textbf{Comprehensive Model Evaluation}: We conduct a systematic evaluation of a diverse set of VLMs and LLMs, including general-purpose models, reasoning-enhanced models, and those supporting multi-image inputs, providing unprecedented comparative insights into their capabilities and limitations.
  \item \textbf{In-Depth and Fine-grained Analysis}: Through detailed statistical and ablation studies, we identify critical factors affecting model performance on juxtaposition-based humor and categorize specific failure patterns.  
  \item \textbf{Practical Model Improvement Strategy}: 
  We propose simple yet effective approaches to improve VLMs' understanding of juxtaposition-based humorous images, offering a practical direction for future research.
\end{itemize}

These contributions collectively establish a rigorous benchmark for evaluating machine humor comprehension while advancing the development of VLMs capable of deeper semantic reasoning in multimodal narratives.

%% file: related_work.tex
\section{Related Work}

\subsection{Vision-Language Model Evaluations}

Recent advancements in large vision-language models (VLMs) have demonstrated remarkable capabilities, including following human instructions and performing tasks such as image captioning, visual question answering, and multimodal reasoning through zero-shot prompting~\cite{ouyang2022training,llama3modelcard,minaee2024large,yin2023survey,radford2021learning}. To systematically assess these models, numerous benchmarks have been developed for both language-only~\cite{zheng2024judging,dubois2024alpacafarm,wang2023pandalm,huang2024c} and vision-language~\cite{ying2024mmt,bitton2023visit,bitton2023breaking,li2023seeda,li2023seedb} tasks. However, despite their effectiveness in measuring fundamental abilities such as linguistic comprehension~\cite{parcalabescu2021valse,oh2024exploring} and general problem-solving~\cite{roberts2024smart}, existing evaluations often overlook deeper aspects of contextual reasoning and social intelligence. 
This limitation is critical, as VLMs still struggle to engage in nuanced social reasoning and accurately interpret human-centric contexts~\cite{hu2024viva,li2023oxfordtvg}. Without rigorous assessments of these advanced capabilities, AI systems risk misinterpretations in real-world applications, where nonlinear inference and social considerations are crucial.
Unlike existing works, we propose the development of targeted evaluation tasks to rigorously assess complex semantic reasoning and multimodal situational understanding. Such enhanced evaluations will not only diagnose the strengths and limitations of current models but also provide clear directions for advancing socially aware AI.




\subsection{Computational Humor}
Humor is a fundamental aspect of human communication, and its computational understanding has become a growing area of research~\cite{palmer2003taking,de2016brain}. Early studies primarily focused on textual humor, leveraging linguistic features such as wordplay, incongruity, and sentiment shifts to detect~\cite{chen2017predicting,cattle2018recognizing,yang2015humor} and generate humor~\cite{amin-burghardt-2020-survey}. While these approaches advanced humor detection, even state-of-the-art large language models (LLMs) like ChatGPT still struggle with nuanced humor comprehension~\cite{jentzsch2023chatgpt,zhong2024let}, particularly in cases requiring deep narrative understanding. Although LLMs can recognize humor to some extent, they often fail to grasp the intricate, context-dependent humor that arises in complex narratives.

More recently, research has expanded into multimodal humor analysis, integrating visual and textual elements to predict humor across different formats. Studies on humorous cartoon captions~\cite{shahaf2015inside,hessel-etal-2023-androids,radev2015humor}, visual humor prediction~\cite{jain2024ai,chandrasekaran2016we}, and humor detection in videos~\cite{kayatani2021laughing,liu2024comment} have shown promising advancements. Similar to our work, prior research~\cite{hessel-etal-2023-androids, yang2024can} has explored AI-driven comprehension of memes~\cite{hwang-shwartz-2023-memecap}, cartoons~\cite{radev-etal-2016-humor}, and comics~\cite{hessel-etal-2023-androids,yang2024can,wang2025beyond}, primarily focusing on humor within isolated images. However, a notable gap in the literature is the insufficient exploration of humor embedded within multi-panel narratives.

Unlike single-panel humor, which often relies on direct punchlines, multi-panel comics involve sequential storytelling, evolving contradictions, and juxtaposition to construct humor. These characteristics contribute to deeper and more complex humor structures. Our work leverages these features to investigate how comic juxtapositions—characterized by contradictory narratives—challenge existing computational models. 



\begin{figure*}[!t]
    \centering
    \includegraphics[scale=0.8]{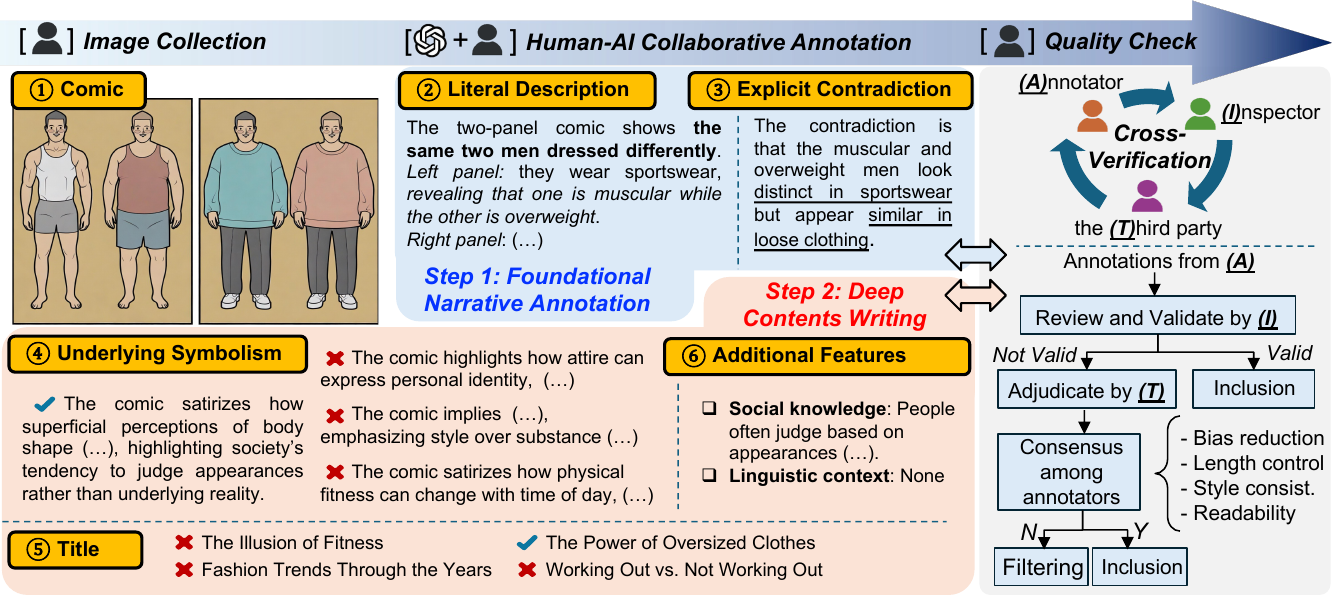}
    \caption{Overview of the Data Construction Pipeline. The dataset construction begins with manually \textit{collecting images} from social media platforms, verified by human reviewers to ensure authenticity and relevance. Next, a \textit{progressive human-AI collaborative annotation} stage is employed to enhance labeling accuracy and efficiency. Finally, a rigorous \textit{quality control and cross-verification} stage is conducted with multiple annotators to refine and validate the dataset.
    }
    \label{fig:annotation_pipeline}
\end{figure*}

\subsection{Visual Reasoning}
Recent research has leveraged neurosymbolic methods~\cite{chen2023see, ke2024hydra, gupta2023visual} to enhance image understanding. These methods typically break down visual reasoning into explicit steps, using powerful vision models to improve object recognition. Another approach, such as Vision-Language Models (VLMs)~\cite{Qwen-VL,wang2023cogvlm, ye2023mplugowl2,achiam2023gpt,liu2024llavanext}, focuses on strengthening the vision encoder and improving alignment through visual prompt tuning. While these techniques have made significant progress in surface-level image understanding, our task requires a deeper level of visual reasoning. Specifically, our goal is to detect and comprehend contradictions between two comic panels—something that goes beyond suffacial image understanding.

Previous research has explored the visual reasoning capabilities of large models in various tasks, including commonsense reasoning~\cite{wang2023gemini, zellers2019recognition, bitton2023breaking}, visual question answering~\cite{hudson2019gqa}, visio-linguistic compositionality~\cite{thrush2022winoground}, and science question answering~\cite{lu2022learn}. However, our work shifts the focus to comparative reasoning with nonlinear narrative. Unlike linear narrative, which follows structured rules and is relatively straightforward for AI to process, comparative reasoning requires models to interpret ambiguous, multi-layered information without explicit guidance. This makes the task significantly more challenging, as it demands a deeper level of natural language understanding and cognitive modeling.

Our work is also relevant to comparative reasoning. Previous studies have primarily focused on textual scenarios~\cite{yu-etal-2023-pre,jindal2006identifying}. Recently, this task has been extended to multimodal contexts involving image pairs\cite{kil2024mllm}. Our task, however, presents a unique challenge: the two panel images in comics collectively construct a contradictory narrative. This requires models to perform both local comparisons between panel elements and global narrative interpretation to grasp the comic's underlying message. Unlike previous approaches, our task demands not only comparative reasoning between visual elements but also an understanding of human behavior and social dynamics to accurately interpret the narrative's intended meaning.


%% file: dataset.tex
\section{The \yesbut (V2) Dataset}

Our benchmark consists of comics with juxtaposed panels that feature an inherent contradiction or unexpected narrative twist. Specifically, each sample includes (1) a two-panel comic that contains a contradictory narrative; (2) a literal description of the scene; (3) an explicit contradiction statement that clarifies the humorous contrast; (4) the underlying symbolism or message conveyed by the comic; (5) a title summarizing the overall theme; and (6) additional features, including social knowledge and linguistic context necessary for interpreting the comic.

Different from the previous work~\cite{hu2024yesbut}, we incorporate additional background features as a fundamental component of our dataset, as it plays a crucial role in comic interpretation. Social knowledge refer to human norms, cultural contexts, and awareness of social events and references 
that are necessary to interpret the comic’s meaning. Linguistic context refers to the language and cultural background in which the comic was created, influencing how the narrative is perceived. Our dataset includes images in multiple languages (\eg English, Chinese, and Russian), with variations in linguistic and cultural environments affecting language comprehension. 


The process of data construction is introduced in Section~\ref{sec:DataConstruction}, followed by a summary of key statistics in Section~\ref{sec:DataStatistics}. Based on annotated components, we designed multi-tiered tasks to systematically evaluate comic comprehension (Section~\ref{sec:TaskDesign}).

\subsection{Dataset Construction}
\label{sec:DataConstruction}
To construct the dataset, we first introduce the \textit{image collection} process, followed by a two-stage annotation process: a \textit{progressive human-AI collaborative annotation} stage and a \textit{quality check and cross-verification} stage. Fig.~\ref{fig:annotation_pipeline} provides an overview of the data construction pipeline.

\subsubsection{Image Collection} 
Our dataset consists of captionless comics collected from social media platforms\footnote{\url{https://twitter.com} and \url{https://www.pinterest.com/}}. These comics depict conflicting narratives of everyday life, capturing humorous, ironic, or thought-provoking scenarios. To ensure data quality and relevance, we applied several preprocessing steps. First, we implemented deduplication techniques to remove identical or near-duplicate images, enhancing dataset diversity. Next, we filtered out images with ambiguous or unclear meanings to minimize noise and facilitate meaningful analysis. Additionally, we conducted content moderation to exclude comics containing inappropriate, offensive, or harmful material. After completing these preprocessing steps, our final curated dataset comprises 1,264 captionless comics, offering a valuable resource for studying visual storytelling, humor, and contrasting perspectives in everyday life.

\subsubsection{Progressive Human-AI Collaborative Annotation} 
For each comic, we annotate the five key components: literal description, contradiction explanation, underlying symbolism, comic title, and relevant background knowledge (\ie social norms, linguistic context, and whether it contains text). To ensure high-quality annotations, we primarily rely on human annotators to provide gold-standard labels. The annotation process involved eleven human judges, who were selected based on linguistic and cultural proficiency. They underwent a brief training session to familiarize themselves with the annotation guidelines and ensure consistency across judgments. Throughout the process, annotators worked independently, and disagreements were resolved through discussion. 

To minimize the cost and effort of manual annotation while ensuring high-quality data, we developed a progressive human-AI collaborative pipeline that leverages GPT-4 for structured data annotation. This pipeline functions as an interactive dialogue system, where AI and human annotators collaborate iteratively to refine and enhance annotations across multiple steps. The annotation process begins with AI-assisted generation: given a comic image, GPT-4 initially produces a narrative description along with an explanation of its contradictory logic (\ie \textbf{Step 1: Foundational Narrative Annotation}). These outputs then undergo a collaborative refinement phase, during which human annotators review, correct, and enhance them to establish the gold-standard descriptions and contradictions. Once validated, these refined annotations serve as the foundation for deeper annotations. 

Building upon this initial step, GPT-4 is further prompted to generate deeper and progressively complex insights, including underlying symbolism, a compelling comic title, and relevant background knowledge (\ie \textbf{Step 2: Deep Contents Writing}). At each step, human annotators actively engage with AI-generated content, verifying its accuracy, contextual appropriateness, and interpretative depth. Through this recursive collaboration, the pipeline ensures that AI outputs are not only computationally efficient but also aligned with human-level understanding and nuance.

Structuring the annotation process into progressive steps allows AI to tackle increasingly complex tasks while reducing cognitive load and maintaining high interpretative accuracy. For instance, the pipeline first generates simpler, literal narratives, which serve as a foundation for interpreting the deeper symbolic meanings in comics. This approach enhances efficiency and reduces annotation costs.
Additionally, for the underlying symbolism and title generation tasks, GPT-4 generates hard negative counterparts and distractions to construct multiple-choice questions for our experiments. Human annotators also annotate panel bounding boxes, which are used for panel separation in experiments and included in the dataset. Example prompts used for annotation are provided in Appendix~\ref{sec:appendix_exp_rompts}.




\begin{table}[t!]
\begin{center}
\caption{ Data Statistics. Avg. Len. is the average number of words.
}
\label{tab:avg_word_count}
 \resizebox{0.8\linewidth}{!}{
\begin{tabular}{rr|cc}
\toprule
\multicolumn{2}{r|}{\textbf{\yesbut Components}} &\textbf{\#Num} & \textbf{Ave. Len.} \\ 
\midrule 
\multicolumn{2}{r|}{Image}                & 1,262 & -                    \\ 
\multicolumn{2}{r|}{Literal Description}  & 1,262 & 134                  \\ 
\multicolumn{2}{r|}{Explicit Contradiction} & 1,262 & 33                 \\ 
\multicolumn{2}{r|}{Underlying Symbolism}   & 5,048 & 26                  \\
\multicolumn{2}{r|}{Title}                  & 5,048 & 6                   \\ 
\midrule
\multicolumn{1}{r|}{\multirow{2}{*}{\begin{tabular}[c]{@{}c@{}}Addition \\ Features\end{tabular}}} & Social Knowledge &3,407  & 97 \\
\multicolumn{1}{r|}{}    & Linguistic Context & 1,262&  1\\

\bottomrule
\end{tabular}
}

\end{center}
\end{table}

\subsubsection{Quality Check with Cross-Verification} 

The annotation process incorporates multiple quality checks to ensure accuracy, consistency, and consensus among annotators. We implemented a cross-verification procedure in which each annotated comic undergoes review by a designated inspector, who verifies correctness and flags any ambiguities or quality concerns. If an annotation is unclear or inconsistent, a third annotator serves as an adjudicator to make the final decision. Comics with unresolved ambiguities, controversial content, or potential bias are filtered out. Finally, one of the authors conducts a comprehensive review to validate the annotations before their inclusion in the benchmark.

\begin{figure*}[t]  
    \centering
    \includegraphics[width=\textwidth]{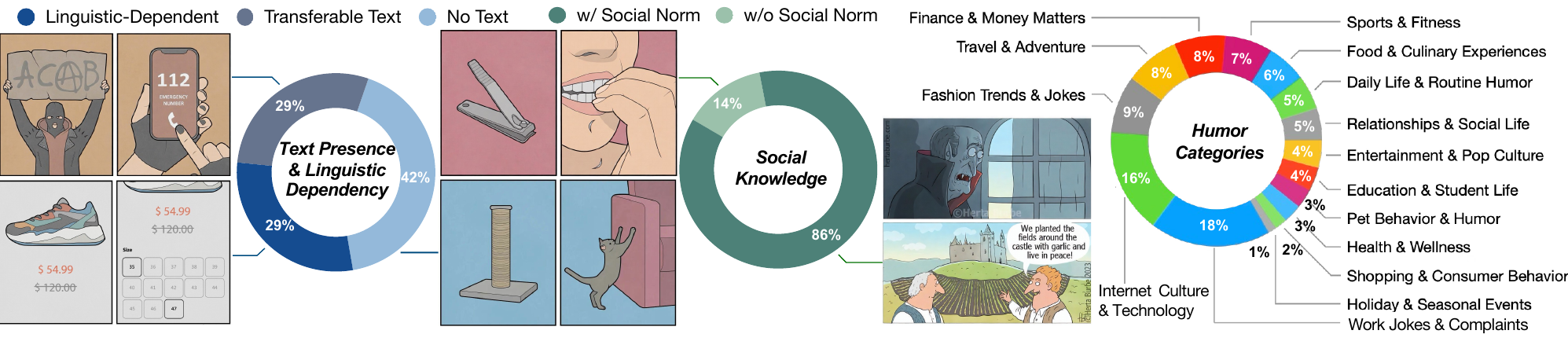}  
    \caption{Distribution of the original 1,264 comics downloaded from social media based on different aspects, including embedded text presence, reliance on social knowledge, and distinct humor categories.
    Overall, we show that our \yesbut exhibits balanced text presence, provides insights into social norms and cultural expectations, and captures a diverse thematic range of humor.
    }
    \label{fig:attributes_distribution}
\end{figure*}

\subsection{Dataset Statistics and Attribute Distribution} 
\label{sec:DataStatistics}
We analyze key attributes in our dataset that influence VLM comprehension of narrative contradictions and humorous contrasts in comics. Our dataset consists of 1,262 comics, each accompanied by high-quality annotations. A statistical breakdown of annotated components, including their quantity and length, is presented in Table~\ref{tab:avg_word_count}.

Beyond basic statistics, we examine the distribution of comics across various categorical attributes to gain deeper insights into the dataset’s structure. Fig.~\ref{fig:attributes_distribution} illustrates these distributions, with representative examples from each category. A comprehensive breakdown of category-wise statistics is provided in Appendix~\ref{sec:appendix_data_distribution}.

\subsubsection{Presence of Text and Linguistic Context}
Our dataset includes both text-embedded and purely visual comics, allowing us to investigate how textual elements influence model performance. To facilitate this analysis, each comic is annotated to indicate the presence or absence of embedded text. This distinction allows for an assessment of whether textual components enhance model comprehension or if models can primarily infer humor from visual cues. Notably, 58\% of the comics in our dataset contain embedded text, while 42\% rely solely on visual elements (Fig.~\ref{fig:attributes_distribution}, left). This balanced distribution enables a comprehensive evaluation of how models process different modalities of humor and the extent to which textual information contributes to their interpretative abilities.

For comics containing embedded text, we further categorized them based on linguistic dependency, recognizing that language variations influence humor perception. The two linguistic categories are defined as follows:
\begin{itemize}
    \item \textbf{Linguistic-Dependent:} Comics that require knowledge of a specific language (e.g., English, Chinese, or Russian) to understand the humor. These typically involve puns, homophones, idioms, or culturally specific wordplay.
    \item \textbf{Transferable Text:} Comics containing text that can be directly translated into another language (\eg, Chinese to English) without altering its humor or meaning. This category also includes comics in which humor arises purely from visual cues, making them universally understandable regardless of language.
\end{itemize}

\subsubsection{Social Knowledge}
Many comics require social knowledge, including human norms, cultural contexts, or common sense beyond simple visual interpretation. To quantify the impact of social knowledge, we provide annotations that highlight comics requiring deeper contextual understanding beyond basic visual or texture elements. These annotations distinguish comics that are self-explanatory and those that demand external social knowledge or cultural norms for interpretation. As illustrated in Fig.~\ref{fig:attributes_distribution} (middle), 86\% images require social knowledge for proper interpretation. Only a small fraction can be readily understood without additional background knowledge.

\subsubsection{Humor categories} 
To systematically analyze humor and encompass its diverse thematic dimensions, we categorize humorous content into 15 domains, each representing distinct contexts and topics frequently encountered by audiences (Fig.~\ref{fig:attributes_distribution}, right). 
The dataset predominantly features humor about Work Jokes \& Complaints (18\%) and Internet Culture \& Technology (16\%), highlighting workplace frustrations and digital-era satire. 
Mid-tier categories include Fashion Trends \& Jokes (9\%), Travel \& Adventure (8\%), Finance \& Money Matters (8\%), and Sports \& Fitness (7\%), emphasizing humor derived from personal experiences, social trends, and lifestyle habits. 
Additional everyday-life categories ($\leq$ 6\%) encompass Food \& Culinary Experiences, Daily Life \& Routine Humor, Relationships \& Social Life, Entertainment \& Pop Culture, and Education \& Student Life. 
Lastly, niche categories ($\leq$ 3\%) like Pet Behavior \& Humor, Health \& Wellness, Shopping \& Consumer Behavior, and Holiday \& Seasonal Events, highlight humor derived from specific cultural contexts and consumer behaviors. It can be seen that the collected comics cover a broad thematic scope and its reflection of social, cultural, and individual experiences.


\subsection{Task Design: Do Large Models Understand Humor in Juxtaposition?}
\label{sec:TaskDesign}

Our goal is to assess the ability of contemporary large-scale (visual) language models to recognize and interpret humor arising from narrative contradictions commonly found in comics. 
To achieve this, we have devised four targeted tasks that progressively assess the model’s comprehension, reasoning abilities—particularly non-linear reasoning—and sensitivity to the nuanced, abstract, and symbolic content embedded in comic narratives:

\noindent\textbf{Task 1: Literal Description Writing.} 
This initial task measures the model's basic narrative comprehension. Given a two-panel comic as input, the model is tasked with generating a concise textual description capturing the overarching narrative depicted across the panels. Unlike standard image captioning tasks, which focus on isolated visual details, this exercise requires the model to synthesize the events into a coherent textual narrative.

\noindent\textbf{Task 2: Explicit Contradiction Generation. }
Moving beyond literal interpretation, this task evaluates the model's ability to recognize the contradictions arising from the juxtaposed panel of the comics. Presented again with the comic as input, the model must generate a textual explanation clearly identifying and elaborating on the contradiction embedded within the narrative. Successful performance indicates the model's capability to reason logically and non-linearly about narrative inharmoniousness, which is crucial for humor understanding.

\noindent\textbf{Tasks 3: Underlying Symbolism Selection.} Humor often conveys deeper symbolic or conceptual messages beyond its surface narratives. In this task, we evaluate the model’s sensitivity to these implicit symbolic meanings. The model is provided with a comic and four possible symbolic interpretations—one correct and three carefully designed distractors that appear plausible but are incorrect. By selecting the correct interpretation, the model demonstrates its ability for abstract reasoning and deep narrative comprehension.

\noindent\textbf{Task 4: Title Matching.} 
This final task evaluates whether the model can associate comics with an appropriate title. Given a comic strip and four possible titles (one correct and three plausible distractors), the model selects the most suitable title summarizing the underlying meaning and narrative abstraction of the comic. Because the title encapsulates the nuanced, abstract humor within the narrative, successful performance indicates a sophisticated grasp of both humor and high-level narrative understanding.

Collectively, these tasks provide a robust evaluation framework to explore multiple levels of humor comprehension, ranging from surface literal and logical interpretations to in-depth abstract symbolic reasoning, ultimately revealing whether large-scale language models genuinely understand humor in juxtaposition.

%% file: experiment.tex
\section{Experiments}

\begin{table*}[t]
    \centering
    \caption{Model performance on tasks ranging from foundational narrative understanding to deep content reasoning. Our evaluation includes a diverse set of VLMs: general-purpose models, reasoning-enhanced models, and models capable of processing multi-image inputs. Additionally, we assess advanced LLMs, which use captions generated by LLaVa-Next-13B as input. 
    For the tasks of literal description and contradiction generation, we report the BERT score (F1), ROUGE-2 (F1), and GPT evaluation scores. The best scores are highlighted in \textbf{bold}, while the second-best scores are \underline{underlined}. }
    \resizebox{0.98\textwidth}{!}{
    \begin{tabular}{@{}rl cccccccc@{}}
        \toprule
        \multirow{3}{*}{\textbf{Type}} & \multirow{3}{*}{\textbf{Model}} & \multicolumn{3}{c}{\textbf{Literal Description}} & \multicolumn{3}{c}{\textbf{Contradiction}} & \textbf{Symbolism} & \textbf{Title} \\
        \cmidrule(l){3-5} \cmidrule(l){6-8} \cmidrule(l){9-9} \cmidrule(l){10-10} 
        & & BERT & R-2 & GPT & BERT & R-2 & GPT & Accuracy & Accuracy \\
        
        \midrule
        \multirow{4}{*}{\begin{tabular}[c]{@{}r@{}}General-Purpose\\ VLMs\end{tabular}} & LLaVA-1.5-7B & 86.67 & 59.31 & 3.32 & 85.97 & \underline{57.04} & 3.23 & 51.13 & 64.50 \\
         & LLaVA-1.5-13B & 85.62 & 55.97 & 3.31 & 85.86 & 56.18 & 3.20 & 71.57  & 68.31\\
         & CogVLM2 & 86.58 & 56.05 & 3.37 & 86.58 & 56.05 & 2.98 & 37.40 & 54.43 \\
         & GPT4-Vision-Turbo  & 87.41 & 60.02 & 3.65 & 87.27 & 47.98 & 3.55 & 74.05 & 71.36 \\
         
        \midrule
         \multirow{5}{*}{\begin{tabular}[c]{@{}r@{}}Multi-Image\\ VLMs\end{tabular}} & LLaVA-OneVision-0.5B & 85.15 & 46.57 & 3.11 & 86.99 & 43.20 & 2.22 & 36.87 & 38.52 \\
         & LLaVA-OneVision-7B & 85.26 & 42.13 & 3.55 & 87.63 & 46.84 & 3.02 & 67.64 & 70.80 \\
         & LLaVA-OneVision-72B & 86.42 & 52.03 & 3.63 & 87.61 & 46.82 & 3.55 & \underline{80.30} & 78.48 \\
         & Qwen2-VL-7B & 87.16 & 57.10 & 3.55 & 86.83 & 55.70 & 3.09 & 74.48 & 74.72 \\
         & Qwen2-VL-72B  & \underline{88.16} & \underline{60.84} & 3.77 & \underline{87.71} & \textbf{60.07} & 3.49 & 79.98 & \textbf{81.25} \\
         
         \midrule
        \multirow{4}{*}{\begin{tabular}[c]{@{}r@{}}Reasoning-\\Enhanced VLMs\end{tabular}} & LLaVA-Next-7B & 86.12 & 57.33 & 2.94 & 85.55 & 56.05 & 2.57 & 59.35 & 56.57 \\
         & LLaVA-Next-13B & 86.49 & 56.92 & 3.05 & 85.57 & 55.83 & 3.12 & 70.36 & 66.88 \\
         & LLaVA-Next-72B & 86.00 & 56.00 & 3.26 & 87.15 & 46.23 & 3.31 & 74.72 & 71.63 \\
         & GPT-4o & \textbf{88.96} & \textbf{63.13} & \textbf{3.96} & \textbf{87.85} & 49.07 & \textbf{3.72} & \textbf{80.38} & \underline{80.62} \\
        
         \midrule
         \multirow{6}{*}{LLMs} 
         & GPT-4 & - & - & - & 86.57 & 46.32 & 2.86 & 61.85 & 59.89 \\
         & Deepseek-r1-70B & - & - & - & 87.20 & 45.38 & 3.37 & 65.32 & 57.47 \\
         & Llama3-8B & - & - & - & 86.62 & 43.95 & 3.24 & 60.62 & 55.71 \\
         & Llama3-70B & - & - & - & 86.71 & 43.12 & 3.52 & 67.35 & 62.68 \\
         & Qwen2.5-7B & - & - & - & 86.60 & 46.20 & 3.28 & 64.98 & 56.58 \\
         & Qwen2.5-72B & - & - & - & 86.74 & 46.33 & 3.31 & 67.59 & 71.47 \\
        \bottomrule
    \label{tab:main_results}
    \end{tabular}
    }

\end{table*}

This section outlines the models, evaluation metrics, and settings used in our experiments. The goal is to evaluate model performance across different tasks, including text generation (literal description writing and contradiction generation) and multiple-choice questions (symbolism selection and title matching). We consider two types of models: Vision-Language Models (VLMs), which process both images and text, and Large Language Models (LLMs), which rely on image-to-text conversion before evaluation.

\subsection{Models and Settings}
We evaluate a diverse set of VLMs and LLMs in a zero-shot setting.
For VLMs, we consider three categories:
(1) General-purpose VLMs that support single-image input, including LLaVA-1.5~\cite{liu2023improved}, CogVLM~\cite{hong2024cogvlm2}, and GPT-4v~\cite{achiam2023gpt};
(2) Multi-image finetuned models that allow processing of multiple images as input, including LLaVA-OneVision~\cite{li2024llava}, Qwen2-VL~\cite{wang2024qwen2}, and ChatGPT4-Vision-Turbo~\cite{achiam2023gpt}; 
(3) Reasoning-Enhanced VLMs specifically designed to improve reasoning capabilities, such as LLaVA-Next~\cite{liu2024llavanext} and GPT-4o~\cite{hurst2024gpt}.

For LLMs, we evaluate GPT-4~\cite{zhu2022minigpt4}, Deepseek-r1~\cite{guo2025deepseek}, Llama3~\cite{llama3modelcard} and Qwen2~\cite{yang2024qwen2}. Since LLMs cannot process images directly, we first use LLaVA-1.6 13B~\cite{liu2024llavanext} to generate captions that serve as literal descriptions for each comic~\footnote{We select LLaVA-1.6 specifically for its strong image captioning capabilities and because its parameter count is comparable to the LLMs.}. These captions replace the original images, allowing the LLMs to process them alongside the questions. As a result, LLMs are not evaluated on the literal description task.

For implementation, we set the temperature to 1 for GPT-4o and ChatGPT. For all other models, we maintain default parameter settings and prompt templates during inference. To mitigate prompt variance across different tasks, we create three distinct prompts per task and report the average score across these runs. Further details on model specifications, prompt templates, and experimental procedures are provided in the Appendix~\ref{sec:appendix_eval_prompts}.

\subsection{Evaluation Metrics}
To ensure a thorough assessment, we employ both automated metrics and human evaluations. The automated evaluation uses multiple metrics to capture different aspects of model performance, while human evaluations help assess qualitative aspects of text generation.
For multiple-choice question tasks (\ie underlying symbolism selection and title matching), we adopt accuracy as the primary evaluation metric. For text generation tasks (\ie literal description writing and contradiction generation), we follow text generation research~\cite{celikyilmaz2020evaluation,gao2024llm} and employ a diverse set of metrics: word overlap-based metrics such as ROUGE-2 (R-2), word vector-based metrics including BERT Score, and GPT-based Score. Recent studies have demonstrated that GPT-based evaluation methods align closely with human judgment~\cite{zheng2023judging}. By incorporating multiple metrics, we aim to comprehensively assess the quality of model-generated descriptions and contradiction texts across different dimensions.

Recognizing the limitations of automatic evaluation for text generation tasks, human evaluations are conducted to provide additional insights. We establish a 5-point scoring system (1=worst, 5=best) and engage human judges to evaluate several aspects. For literal description, we evaluate: \textit{Correctness}, which measures accuracy in conveying the comic's narrative; \textit{Completeness}, which assesses coverage of all important narrative elements; and \textit{Fidelity}, which examines the absence of hallucinations, ensuring all content is supportable by the comic images. For contradiction generation, we evaluate on aspects of Correctness and Fidelity using the same criteria. Please refer to Appendix~\ref{sec:appendix_human_eval} for more details.



%% file: results.tex
\section{Main Results}

In this section, we present a comprehensive analysis of model performance across the tasks. Table~\ref{tab:main_results} shows the main results for all models tested. 

\subsection{Literal Description Writing} 
For literal description writing, the results show that commercial models generally outperform open source alternatives. Among all  models, GPT-4o achieves the highest performance scores. For open source models with identical architectures, we observe a positive correlation between model size and performance, suggesting that larger models possess enhanced comic comprehension and description generation capabilities. Notably, Qwen2-VL-72B ranks second, with performance metrics approaching those of GPT-4o, indicating a narrowing gap between commercial and open source VLMs.


\subsection{Contradiction Generation}

For the contradiction generation task, we include both VLMs and LLMs, with the latter utilizing image captions generated by LLaVA-1.6 13B as input. Our evaluation employing multiple metrics reveals that GPT-4o achieves superior performance in BERT and GPT scores, indicating high semantic similarity and language quality, while Qwen2-VL demonstrates the highest ROUGE-2 (R-2) score, suggesting better lexical overlap with reference contradictions. Interestingly, GPT-4o's ROUGE-2 score was surpassed by several 7B-parameter models despite its overall stronger performance. This discrepancy can be attributed to the inherent limitations of ROUGE-2 evaluation, which primarily measures bigram overlap. Upon manual analysis of the generated outputs, we observed that GPT-4o typically produces longer outputs with paraphrasing and expanded context, which results in reduced lexical overlap with reference contradictions.

Moreover, we can observe that the LLaVA-OneVision series models consistently outperforms both LLaVA-1.5 and LLaVA-Next series models in this task. We hypothesize that this superior performance stems from their multi-image training approach, which enhances their ability to detect and reason about relationship changes across multiple panels—a crucial skill for understanding comic narratives and accurately capturing inter-panel relationships.


\subsection{Deep Reasoning Tasks}
\label{sec:res_deep_reasoning}

The underlying philosophy selection and title matching tasks require sophisticated reasoning based on comic narratives. Our results show that for the underlying symbolism selection task, GPT-4o achieves the highest accuracy (80.38\%), while for the title matching task, Qwen2-VL-72B demonstrates the highest accuracy of 81.25\%.

A key finding is that larger models generally demonstrate better comic comprehension capabilities, aligning with previous research indicating enhanced reasoning abilities in models with higher parameter counts. Comparing across the LLaVA model series with equivalent parameter counts, LLaVA-Next models consistently outperform LLaVA-1.5 models, while LLaVA-OneVision models generally surpass LLaVA-Next models. This performance hierarchy can be attributed to the progressive improvements in each model series: LLaVA-Next enhances reasoning abilities and world knowledge~\cite{liu2024llavanext}, while LLaVA-OneVision further incorporates supervised fine-tuning on multi-image and video inputs, strengthening its capacity to understand relationships and changes across multiple images. These findings suggest a promising direction for improving VLMs' understanding of juxtaposed humorous comics: enhancing models' reasoning and multi-image relationship comprehension through targeted fine-tuning.

Additionally, we observe that LLMs consistently underperform compared to VLMs of equivalent scale. This performance gap can be attributed to the LLMs receiving literal descriptions generated by LLaVA-Next-13B as input, thus inheriting any errors, information loss, or hallucinations present in the VLM-generated descriptions. We provide a more detailed analysis of description quality's impact on LLM performance in Section~\ref{sec:influece_of_captions}.

Another notable observation is that most models perform worse on title matching than on underlying symbolism selection, with Qwen2-VL-72B being the sole exception. This discrepancy likely stems from titles being shorter and more abstract representations of narratives that do not explicitly convey the underlying comic concepts. Consequently, distinguishing between correct titles and distractors requires deeper, more rigorous understanding and reasoning capabilities, presenting a more significant challenge for most models.

\begin{figure*}[t]
    \centering
    \begin{minipage}{0.24\textwidth}
        \centering
        \includegraphics[width=\linewidth, trim={22cm 0 0 0},clip]{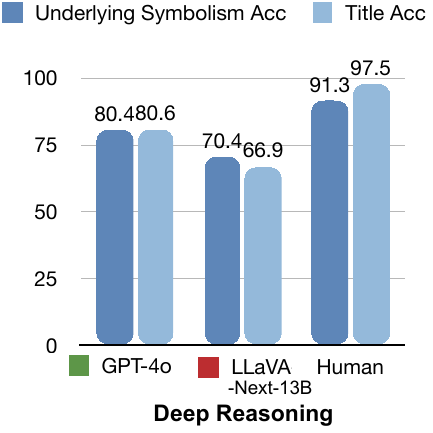}
        \caption{Human performance on deep reasoning tasks.}
        \label{fig:human-eval-deeper-tasks}
    \end{minipage}
    \hfill
    \begin{minipage}{0.74\textwidth}
        \centering
        \includegraphics[width=\linewidth, trim={0 0 7.4cm 0},clip]{images/human_evaluation_scores.pdf} 
        \caption{Human Evaluation of Literal Description and Contradiction Generation Tasks.}
        \label{fig:human-eval}
    \end{minipage}
    
    \label{fig:combined}
\end{figure*}

\subsection{Comparison with Human Performance}

To establish a human performance baseline for better understanding model performance, we conduct a controlled evaluation with three human participants on a sample of 50 randomly selected comics. Participants are asked to complete the same Underlying Symbolism and Title Matching tasks as our evaluated models.


As shown in Fig.~\ref{fig:human-eval-deeper-tasks}, human evaluators substantially outperform even the strongest AI models on both deep reasoning tasks, especially on the more complex title matching task.
This performance gap highlights the significant challenges AI systems face when tasks require complex non-linear reasoning, abstract concept interpretation, and cultural context understanding. These results confirm that despite recent advances in multimodal capabilities, substantial improvements are still needed for VLMs to achieve human-like comprehension of juxtaposition-based humor in comics.

\subsection{Human Evaluation}



To complement our automatic evaluation metrics, we conduct human evaluations on a randomly selected subset of 30 samples to assess the quality of both literal descriptions and contradiction generations. The results, presented in Fig.~\ref{fig:human-eval}, align with the trends observed in automatic evaluations: commercial models consistently outperform open-source alternatives across both tasks, with GPT-4o achieving the highest overall scores. Among open-source models, Qwen2-VL-72B stands out, demonstrating performance levels approaching those of GPT-4o.

Additionally, we observe that models generally achieve higher scores for literal descriptions than for contradiction generation. This performance gap highlights the inherently greater complexity of the contradiction generation task, which requires sophisticated comparative reasoning across multiple visual elements in paired panels. In contrast, generating literal descriptions is a more straightforward process, primarily relying on visual perception of the comic image.

Furthermore, the results indicate a correlation between model scale and performance quality, suggesting that larger parameter counts contribute to improved image comprehension. Notably, models specifically designed for multi-image processing and enhanced reasoning—such as those in the LLaVA-OneVision series—outperform general-purpose VLMs of similar scale, reinforcing our earlier findings.

Lastly, we identify a strong positive correlation between performance on literal description and contradiction generation tasks. Models that excel in one task tend to perform well in the other, indicating that these abilities are fundamental to visual narrative comprehension. Unlike superficial content understanding, interpreting juxtaposed comic panels requires an integrated set of advanced capabilities, including precise image perception, contextual narrative interpretation, extensive world knowledge, and sophisticated reasoning. This interdependence underscores the importance of these core competencies in effective visual understanding.

%% file: Analysis.tex
\section{Analysis and Discussion}
\label{sec:analysis}


In this section, we conduct a series of experiments to investigate various factors influencing the deep reasoning performance of VLMs on comic understanding. Our investigation follows a structured progression:
\textbf{(\textit{A}) Basic Factors Affecting Model Performance}, examining elements such as embedded text and surface descriptions;
\textbf{(\textit{B}) Methodological Enhancements}, incorporating techniques like task decomposition, panel splitting, and model fine-tuning for deep reasoning tasks;
\textbf{(\textit{C}) Broader Enhancements}, assessing the impact of external social knowledge; 
and \textbf{(\textit{D}) Case Study and Error Analysis}, conducting detailed examinations of specific instances and identifying common errors.

\subsection{Basic Factors Affecting Model Performance}\label{sec:analysis:basic_factor}

\subsubsection{{The Role of Embedded Text in Comic Images}}

In our dataset, 58\% of the comics contain embedded text, while 42\% rely solely on visual elements (as depicted in Fig.~\ref{fig:attributes_distribution} left). Analyzing the impact of this embedded text on model performance reveals notable trends, as the results detailed in Table~\ref{tab:text_ablation}.
For the title matching task, all models exhibit improved performance on images with embedded text, especially on comics with transferable text. This improvement suggests that embedded text provides direct contextual information, aiding VLMs in understanding the general content and augmenting the reasoning about the relationships of the comic panels.

Contrary to expectations, in the underlying symbolism deep reasoning task, models other than the Qwen2 series perform worse on images containing embedded text compared to those without. Given that a portion of the comics contain Chinese text (257 images with embedded Chinese text and 113 with English text) a plausible explanation is that Qwen2 models' proficiency in Chinese contributes to their superior performance~\cite{yang2024qwen2technicalreport}. In contrast, other models may lack robust multilingual capabilities, hindering their performance in deep reasoning tasks. Further discussions on language influence are provided in Appendix~\ref{sec:appendix:language_influence}.

These observations underscore the importance of integrating advanced multilingual understanding capabilities into VLMs to enhance their performance in tasks involving complex visual and textual information.

\begin{table}[t]
    \centering
    \caption{Impact of embedded text in comic images on model performance. ``w/o text'' refers to comic images without embedded text. Images containing embedded text, which can assist in comic comprehension, are further categorized into two groups: ling. dept. (Linguistic-Dependent) and trans. txt. (Transferable Text).}
    \resizebox{\columnwidth}{!}{
    \begin{tabular}{lcccc}
        \toprule
        \textbf{Model} & {\textbf{Emb. Text}} & \textbf{Symbolism Acc.} & \textbf{Title Acc.} \\
        \midrule
        {GPT-4o}  & \textit{w/o text} & \textbf{82.36} & 77.42 \\
                 & \textit{ling. dept.} & 77.85 & 77.26 \\
                 & \textit{trans. txt.}  &80.06 & \textbf{88.73}\\
                                      
        \midrule
        {LLaVA-OneVision-7B}  
                                     & \textit{w/o text} & \textbf{71.11} & 69.61 \\
                    & \textit{ling. dept.} & 61.02 &63.80\\
                 & \textit{trans. txt.}  & 69.32& \textbf{82.03}\\

        \midrule
        {LLaVA-OneVision-72B}  
                                      & \textit{w/o text} & \textbf{82.05} & 77.42 \\
                    & \textit{ling. dept.} & 77.09 &72.83 \\
                 & \textit{trans. txt.}  & 81.02&\textbf{85.81} \\

        \midrule
        {Qwen2-VL-7B}  
                            & \textit{w/o text} & 70.11 & 69.17 \\
                & \textit{ling. dept.} & {76.97}  & {77.47}\\
                 & \textit{trans. txt.}  &\textbf{78.33} & \textbf{77.97}\\

        \midrule
        {Qwen2-VL-72B}  
                              & \textit{w/o text} & 77.67 & 76.74 \\
                & \textit{ling. dept.} & \textbf{83.37} & {82.49} \\
                 & \textit{trans. txt.}  & {79.89} &\textbf{86.58} \\

        \bottomrule
    \end{tabular}
    }

    \label{tab:text_ablation}
\end{table}

\begin{figure}[t]
    \centering
    \includegraphics[width=1\linewidth]{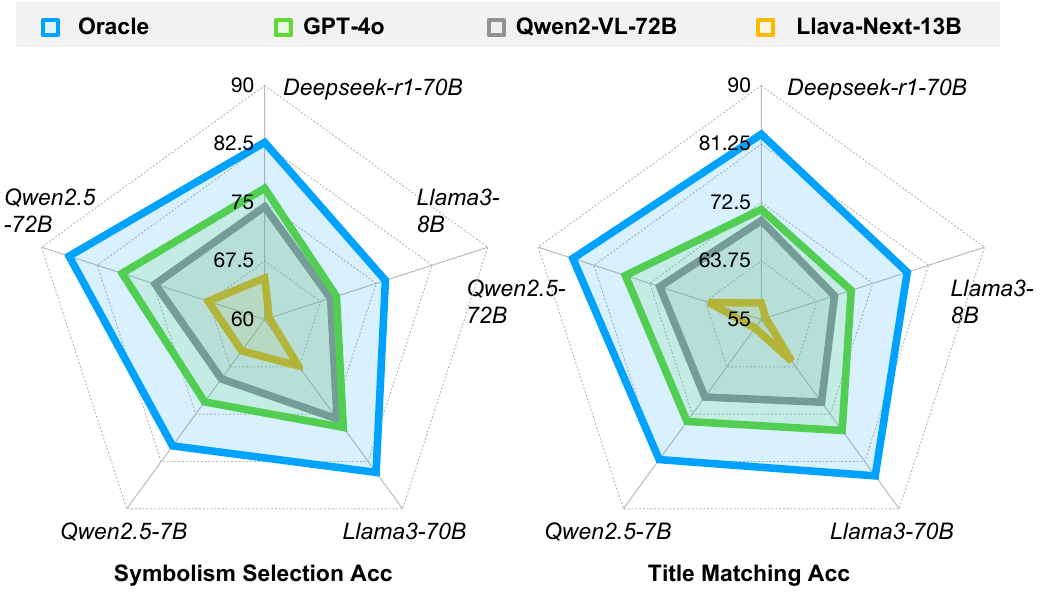}
    \caption{LLMs' performance on deep reasoning tasks using different image descriptions as inputs. These image descriptions are from our annotations (\ie Oracle) or generated from VLMs (\eg GPT-4o, Qwen2-VL-72B, and Llava-Next-13B). }
    \label{fig:llm_captions_res}
\end{figure}

\subsubsection{{Influence of Surface Descriptions on Deep Reasoning}}
\label{sec:influece_of_captions}

Effective deep reasoning about comic narratives fundamentally depends on accurate comprehension of surface content within the images. To investigate this relationship, we examine how the quality of comic descriptions influences subsequent deep reasoning performance across the evaluated models.

For LLMs, which rely entirely on textual descriptions as input for reasoning tasks, we conduct comparative experiments using descriptions generated by different VLMs as well as human-authored oracle descriptions. As shown in Fig.~\ref{fig:llm_captions_res}, descriptions generated by more advanced VLMs (specifically GPT-4o and Qwen2-VL-72B) lead to substantially improved reasoning performance across all evaluated LLMs compared to descriptions from LLaVA-Next-13B. This performance differential is consistent across LLM scales and architectures. Most notably, when provided with human-authored oracle descriptions, all LLMs achieve their highest performance scores, establishing an upper bound on potential performance. 

These findings demonstrate a strong positive correlation between literal description quality and deep reasoning capabilities. The substantial performance variations observed when using identical LLMs with different input descriptions highlight the critical importance of accurate surface-level understanding as a foundation for higher-order reasoning about comic narratives. This suggests that improvements in VLMs' descriptive capabilities could yield cascading benefits for downstream reasoning tasks, even without architectural changes to reasoning components.


\begin{figure}[t]
    \centering
    \includegraphics[trim=0 0 1.2cm 0, clip,width=1\linewidth]{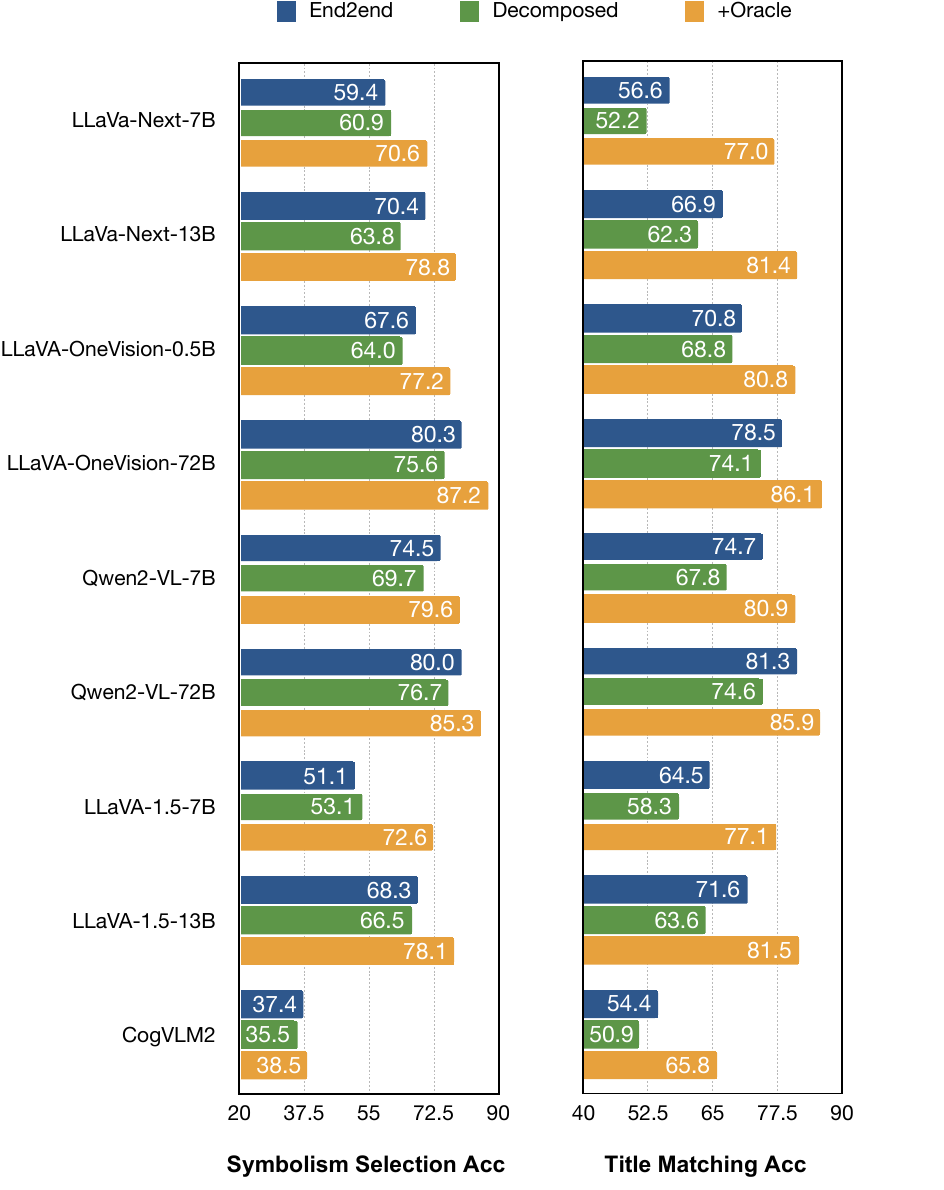}
    \caption{Comparison of VLM deep reasoning performance in end-to-end and decomposed settings. In the decomposed setting, we additionally incorporate oracle literal descriptions provided by humans as an upper bound.}
    \label{fig:decomposed_res_vlm}
\end{figure}

\subsection{Methodological Enhancements}\label{sec:analysis:methodological_enhancements}

\subsubsection{{Does Task Decomposition Lead to Better Deep Reasoning}}
\label{sec:task_decomposition}

VLMs typically approach deep reasoning tasks in an end-to-end manner, simultaneously performing image captioning, narrative understanding, and deep reasoning when presented with comic images. This integrated approach requires models to manage multiple cognitive tasks of varying complexity concurrently. In this section, we investigate whether decomposing the complex reasoning tasks into sequential stages improves performance outcomes.

In particular, we implement a two-stage decomposition approach: first prompting the VLM to generate a literal description of the comic for surface understanding, then directing it to predict the result based on both the original comic image and the generated description. Fig.~\ref{fig:decomposed_res_vlm} presents the comparative results between this decomposed approach and the standard end-to-end reasoning. Contrary to intuitive expectations, our findings indicate that task decomposition with model-generated descriptions does not consistently improve performance. In fact, we observe a performance decline across all models in the title selection task when augmented with their own generated descriptions. However, when we substitute model-generated descriptions with human-authored oracle descriptions in the decomposed setting, all VLMs demonstrate significant performance improvements.

These results suggest several important insights about current VLM capabilities. First, models continue to struggle with accurately recognizing critical visual elements in comic images that are essential for deep narrative reasoning. Second, the introduction of potentially flawed self-generated descriptions can propagate and amplify errors through the reasoning process. Our manual analysis reveals that models frequently generate descriptions containing hallucinations or omit crucial narrative elements (with further discussions in Section~\ref{sec:analysis:case_study}), subsequently misleading their own reasoning processes. This creates a compounding error effect that undermines the potential benefits of task decomposition. These findings highlight the importance of addressing fundamental visual perception and description accuracy as prerequisites for improving complex reasoning capabilities in future VLM development.

\begin{table}[t]
    \centering
    \caption{Comparison of model performance using a \textit{Single} comic image versus \textit{Multi}-images split by panels. For each setting, we present results of both end-to-end and decomposed predictions as in Section~\ref{sec:task_decomposition}.}
    \resizebox{\columnwidth}{!}{
    \begin{tabular}{l l l c c}
        \toprule
        \textbf{Model} & \multicolumn{2}{l}{\textbf{Setting}} & \textbf{Symbolism Acc.} & \textbf{Title Acc.} \\
        \midrule
        
        \multirow{5}{*}{LLaVA-OneVision-0.5B} & \multirow{2}{*}{\textit{Single}} & End2end & 36.87 & 38.53 \\
        & & Decomposed & \textbf{39.16} & \textbf{46.44} \\
        \addlinespace \cdashline{2-5} \addlinespace
        & \multirow{2}{*}{\textit{Multi}} & End2end & 38.59 & 38.51 \\
        & & Decomposed & 35.82 & 40.17 \\
        \midrule
        \addlinespace
        
        \multirow{5}{*}{{LLaVA-OneVision-7B}} & \multirow{2}{*}{\textit{Single}} & End2end & \textbf{67.64} & \textbf{70.83} \\
        & & Decomposed & 64.00 & 68.83 \\
        \addlinespace \cdashline{2-5} \addlinespace
        & \multirow{2}{*}{\textit{Multi}} & End2end & 65.45 & 70.21 \\
        & & Decomposed & 55.55 & 57.05 \\
        \midrule
        \addlinespace
        
        \multirow{5}{*}{LLaVA-OneVision-72B} & \multirow{2}{*}{\textit{Single}} & End2end & \textbf{80.67} & 77.10 \\
        & & Decomposed & 67.59 & 58.56 \\
        \addlinespace \cdashline{2-5} \addlinespace
        & \multirow{2}{*}{\textit{Multi}} & End2end & 80.30 & \textbf{78.48} \\
        & & Decomposed & 75.55 & 74.13 \\
        \midrule
        \addlinespace
        
        \multirow{5}{*}{Qwen2-VL-7B} & \multirow{2}{*}{\textit{Single}} & End2end & \textbf{74.48} & 74.72 \\
        & & Decomposed & 69.73 & 67.75 \\
        \addlinespace \cdashline{2-5} \addlinespace
        & \multirow{2}{*}{\textit{Multi}} & End2end & 72.27 & \textbf{75.42} \\
        & & Decomposed & 64.74 & 62.20 \\
        \midrule
        \addlinespace
        
        \multirow{5}{*}{Qwen2-VL-72B} & \multirow{2}{*}{\textit{Single}} & End2end & \textbf{79.98} & \textbf{81.25} \\
        & & Decomposed & 76.66 & 74.60 \\
        \addlinespace \cdashline{2-5} \addlinespace
        & \multirow{2}{*}{\textit{Multi}} & End2end & 78.29 & 79.71 \\
        & & Decomposed & 71.32 & 63.63 \\
        \bottomrule
    \end{tabular}
    }
    \label{tab:analysis_split_image}
\end{table}

\subsubsection{{Does Splitting Comic Panels into Separate Images Enhance Performance}}

In our standard experimental setup, we evaluate VLMs by providing a single composite comic image containing both panels and requiring models to perform deep reasoning across the entire visual narrative. This approach, however, introduces potential challenges in panel disambiguation, as models must correctly identify and distinguish information from left and right panels to accurately interpret their relationship.

Given that some recent VLM architectures, such as the LLaVA-OneVision series, have been specifically trained on multiple image inputs and video sequences, we conduct analysis to see whether separating comic panels into distinct image inputs can improve performance by better aligning with their training paradigm. This approach theoretically enables models to process each panel individually before integrating information across panels, potentially facilitating more precise relationship modeling.

The results are presented in Table~\ref{tab:analysis_split_image}.
Contrary to expectations, the results show that performance consistently decreases when using split panel inputs compared to single composite images across all evaluated models. 
These results suggest that the sequential processing of separated panels appears to disrupt rather than enhance models' ability to capture cross-panel relationships essential for deep reasoning. Moreover, despite training on multi-image inputs, current models still struggle with the particular cognitive challenge of identifying and reasoning about subtle narrative relationships, contradictions, and thematic connections across sequential panels. This finding highlights a critical gap in current VLMs' cross-image reasoning capabilities, particularly for narratively linked visual content that requires integrative understanding rather than independent processing of each visual element.


\subsubsection{{Model Finetuning for Deep Reasoning Tasks}}


\begin{figure}[t]
    \centering
    \includegraphics[width=1\linewidth]{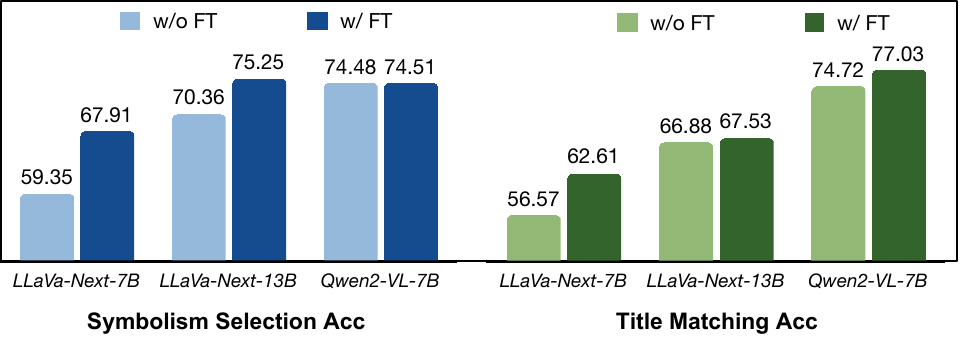}
    \caption{Model performance on deep reasoning tasks with and without fine-tuning. }
    \label{fig:sft_vlms}
\end{figure}

\begin{figure}[t]
    \centering
    \includegraphics[trim=290 230 320 50, clip, width=1\linewidth]{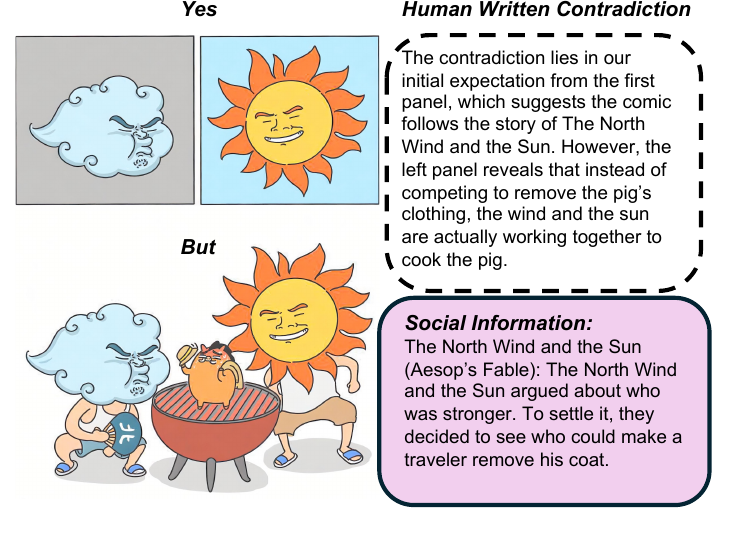}
    \caption{A sample comic that requires additional social knowledge for full comprehension.}
    \label{fig:social_info_example}
\end{figure}

In this section, we explore whether finetuning can enhance model performance on deep reasoning tasks for comic understanding. However, a significant challenge is the scarcity of large-scale comic datasets suitable for training. To address this limitation, we develop a text-only training methodology that circumvents the need for extensive image resources.

Our approach employs a weakly supervised textual data synthesis pipeline utilizing powerful LLMs such as GPT-4o as a data generator. Rather than requiring paired image-text data, we substitute comic images with textual descriptions of narratives, accompanied by corresponding reasoning questions. This method leverages the sophisticated text generation capabilities of LLMs while eliminating the dependency on visual data for training.

Specifically, for data generation, we carefully select 10 diverse examples from our labeled dataset to serve as few-shot prompts. Using these exemplars, we prompt GPT-4o to synthesize 20,000 contradictory comic scene descriptions. For each synthesized scene description, we further employ GPT-4o to generate corresponding questions targeting underlying symbolism selection and title matching tasks. 

This synthetic dataset is then used to finetune only the language components of the VLMs, leaving their visual perception modules unchanged. This targeted approach allows us to enhance the models' comic understanding and reasoning capabilities while maintaining their original visual processing architecture. More detailed information regarding the model training process and data generation methodology is provided in Appendix~\ref{sec:appendix:finetune}.

Fig.~\ref{fig:sft_vlms} presents the comparative results before and after finetuning. The results demonstrate consistent performance improvements across all evaluated models after finetuning, with particularly notable gains in symbolism accuracy for LLaVA-Next-7B (8.56 percentage points) and LLaVA-Next-13B (4.89 percentage points). These results validate the effectiveness of our text-only training approach for enhancing deep reasoning capabilities in VLMs, even without modifying their visual components.

\subsection{Broader Enhancements with Social Knowledge Augmentation}\label{sec:analysis:methodological_enhancements}

\begin{figure*}[t]
    \centering
    \includegraphics[trim=100 82 100 40, clip, width=1\linewidth]{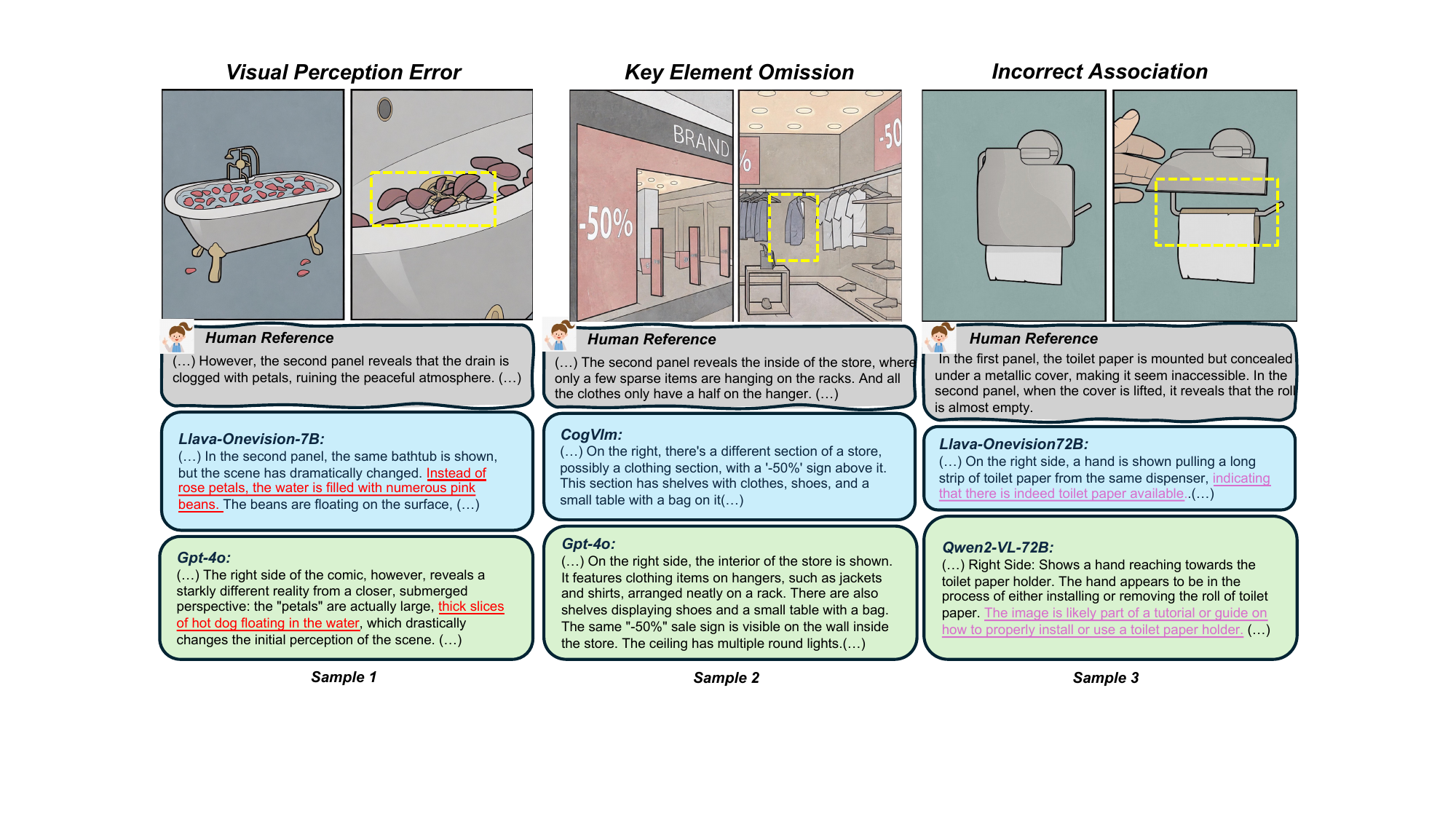}
    \caption{Sample outputs of model-generated literal descriptions with highlighted errors of different types.}
    \label{fig:cases_an}
\end{figure*}

\begin{figure}[t]
    \centering
    \includegraphics[width=1\linewidth]{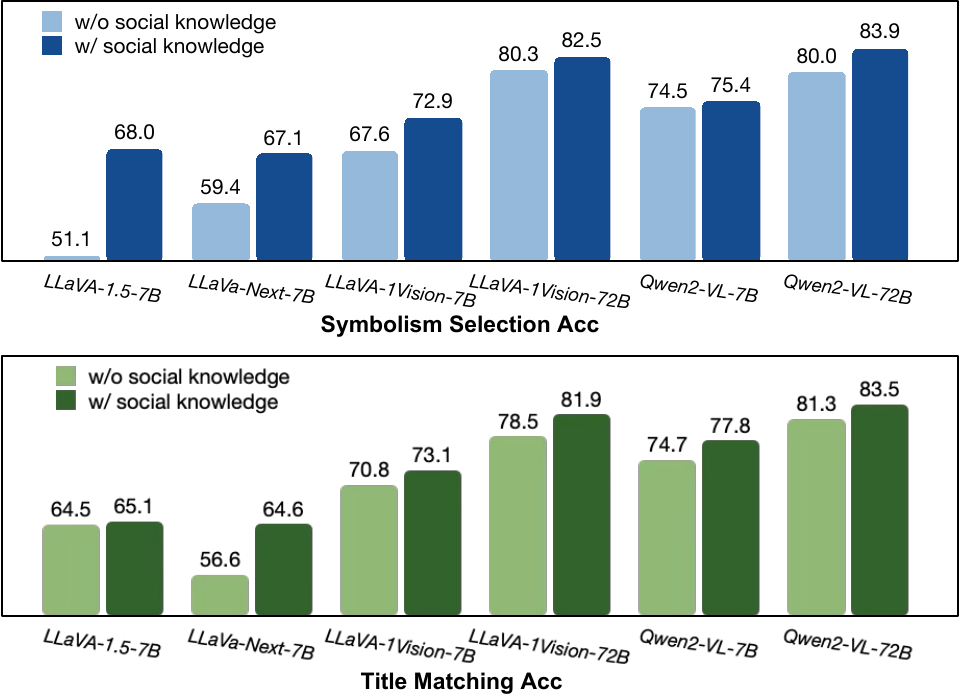}
    \caption{The impact of external social knowledge. }
    \label{fig:knowledge_augmentation}
\end{figure}

Understanding humorous comics fundamentally requires comprehensive knowledge of social events and human behavioral norms (Fig.~\ref{fig:social_info_example}). Prior research has shown that large models continue to exhibit limitations in capturing human intent and social nuances~\cite{hu2023language}. This raises an important question: \textit{Can explicitly integrating social knowledge enhance model comprehension of humor?} To answer this, we designed experiments to test the role of social knowledge in deep reasoning tasks related to comic understanding.

We conduct experiments by supplementing model prompts with our annotated social knowledge specific to each comic's context. As shown in Fig.~\ref{fig:knowledge_augmentation}, all evaluated VLMs achieve either improved or comparable performance when augmented with explicit social knowledge. This consistent pattern of improvement suggests that current models may possess insufficient or inadequately activated social knowledge for effectively interpreting juxtaposed humorous content. The performance gains observed through knowledge augmentation indicate that the models can effectively utilize such information when explicitly provided.

These findings suggest that enhancing models' inherent social knowledge representation or leveraging retrieval mechanisms for relevant knowledge could yield substantial improvements in human-centered reasoning tasks. This points toward promising avenues for developing more socially-aware VLMs through targeted knowledge integration and reasoning enhancement techniques.

\subsection{Case Study and Error Analysis}
\label{sec:analysis:case_study}

To provide qualitative insights into model limitations, we analyze model outputs and present common errors of VLM-generated descriptions in Fig.~\ref{fig:cases_an}. Our analysis reveals several recurring error patterns. We categorize these errors into three types:

\subsubsection{{Visual Perception Error}}
Visual perception error occurs when models incorrectly identify visual elements, attributing erroneous identities or characteristics to objects present in the image. This error type represents a fundamental failure in visual perception. In Sample 1 (Fig.~\ref{fig:cases_an}), we observe that LLaVA-OneVision-7B misidentifies flower petals as ``pink beans,'' while even the more sophisticated GPT-4o incorrectly labels these same petals as ``ham slices.'' These perceptual errors cascade into subsequent reasoning processes, establishing flawed premises that undermine higher-level understanding.

\subsubsection{{Key Element Omission}}

Key element omission errors occur when models fail to recognize or acknowledge significant visual elements present in the comic panels. This issue is particularly common in comics, where the model needs to identify the visual cues to correctly understand the overall narrative. In Sample 2 (Fig.~\ref{fig:cases_an}), which depicts a ``50\% off'' sale showing only half of each clothing item available for purchase (the visual punchline), both CogVLM2 and GPT-4o completely omit this critical visual element in their descriptions. Such omissions eliminate essential information required for understanding the comic's humor.

\subsubsection{{Incorrect Association}}
Incorrect association errors occur when models make up non-existent information or hallucinations for the visual content. This error type often manifests as hallucinated details or narratives that extend beyond what is present in the comic. Sample 3 (Fig.~\ref{fig:cases_an}) demonstrates this error pattern: the comic depicts a nearly-empty paper roll that appears to have more paper than it actually contains. LLaVA-OneVision-72B incorrectly associates this image with an imagined narrative suggesting ``there is still some paper left,'' while Qwen2-VL-72B fabricates a non-existent tutorial context. These hallucinated associations impose incorrect interpretive frameworks that fundamentally alter the comic's intended meaning.

In summary, these error patterns highlight the complex challenges in comic understanding that extend beyond simple visual recognition. Each error type disrupts a different aspect of the interpretive process, from basic perception to contextual framing, collectively undermining models' ability to grasp the nuanced meanings encoded in comic narratives.

%% file: conclusion.tex
\section{Conclusion}
This paper investigates this limitation through the analysis of comics that use juxtaposed panels to create humorous contradictions. We introduce \yesbut, a novel, multi-tiered benchmark designed to evaluate the layered reasoning necessary for humor comprehension, ranging from basic content recognition to deep narrative inference.
Our extensive experiments demonstrate that even state-of-the-art VLMs struggle to match human performance, exposing critical gaps in their ability to grasp nuanced contextual relationships. To address these shortcomings, we propose a novel text-only training strategy that synthesizes textual data to strengthen VLMs' language processing capabilities—eliminating the need for costly image-text paired training data without sacrificing performance.
Beyond identifying key weaknesses in VLMs’ understanding of cultural and creative expressions, our findings chart a promising path toward more robust, context-aware AI models capable of deeper narrative reasoning. This work lays the foundation for improving AI’s ability to process humor, contradictions, and complex multimodal narratives, ultimately advancing human-AI interaction in creative domains.

\section*{Acknowledgments}
All data samples collected are sourced from publicly available content on social media platforms. To maintain content integrity, we carefully review and filter out any samples that may contain offensive or harmful material. The Large Vision-Language Models (VLMs) used in our experiments are pretrained on diverse web corpora, which may inherently introduce biases into their outputs. We encourage users to critically assess the ethical considerations of generated outputs when applying them in future research. Our annotation process involves a team of ten human judges, each compensated with an average hourly wage of \$11, ensuring fair and ethical remuneration for their contributions.

This work made use of the High Performance Computing Resource in the Core Facility for Advanced Research Computing at Case Western Reserve University, which is supported by NSF award NSF-2117439.
We also thank the support from OpenAI Researcher Access grants \#0000007745.

%% file: appendix.tex


\appendices
\clearpage

\section{Ethics Statement}
\label{sec:appendix_ethics_statement}
\subsection{Copyright and License} 
All data samples collected are sourced from publicly available content on social media platforms. We ensure compliance with copyright by utilizing original links to comics without infringement. In addition, we obtained permission from the author artist (e.g., {Anton Gudim, Liz Climo}) to conduct our benchmark using these public images. Additionally, we commit to open-sourcing our annotated benchmark, providing corresponding links to each comic image. We diligently review samples, filtering out potentially offensive or harmful content.

\subsection{Human Annotation} 
Ten human annotators participate in our labeling process, receiving an average hourly wage of \$11 to ensure fair compensation. We take steps to mitigate biases by maintaining a diverse group of annotators and providing clear annotation guidelines.

\subsection{Model Bias \& Ethical Considerations} 
The large vision-language models (VLMs) used in our experiments are pretrained on diverse web corpora, which may introduce biases into their outputs. We encourage users to critically assess potential ethical implications when applying these models in future research.

\begin{figure*}[b]
\Large
\centering

\renewcommand{\arraystretch}{1.8}  
\resizebox{0.9\linewidth}{!}{
\begin{tabular}{c p{18cm}}
\toprule
\textbf{Tasks} & \textbf{Prompts} \\
\midrule
\textbf{Literal Description \& Contradiction} & \textit{The given comic with two panels shows the same situation from two opposite sides with contradictions. You need to first read and understand the comic. Generate a detailed description to illustrate the narrative of the comic and explain the contradiction of what makes the comic interesting or sarcastic.} \\
\midrule
\textbf{Symbolism} & \textit{Write a brief description of the underlying moral of the narrative in one sentence, and include what phenomenon is it satirizing and what we can learn from the comic.} \\
\midrule
\textbf{Title} & \textit{Produce a short eye-catching title reflecting the narrative.} \\
\midrule
\textbf{Negative Symbolism} & \textit{Generate five contextualized, plausible, but ultimately incorrect criticisms and moral lessons we can learn from the image, each in one sentence as distracters. Keep the length and style the same as the correct one.} \\
\midrule
\textbf{Negative Title} & \textit{Provide five seemingly reasonable, eye-catching but incorrect titles.}\\
\midrule
\textbf{Social Knowledge/Norms} & \textit{Based on the image, provide me the necessary social knowledge of the comic.}\\
\bottomrule
\end{tabular}
}
\caption{Prompts for Data Annotation}
\label{fig:annotation_prompts}
\end{figure*}

\section{Data Annotation Details} \label{sec:appendix_data_annotation_details}
\begin{figure*}
    \centering
    \includegraphics[width=0.9\linewidth,trim=90 10 90 10, clip,]{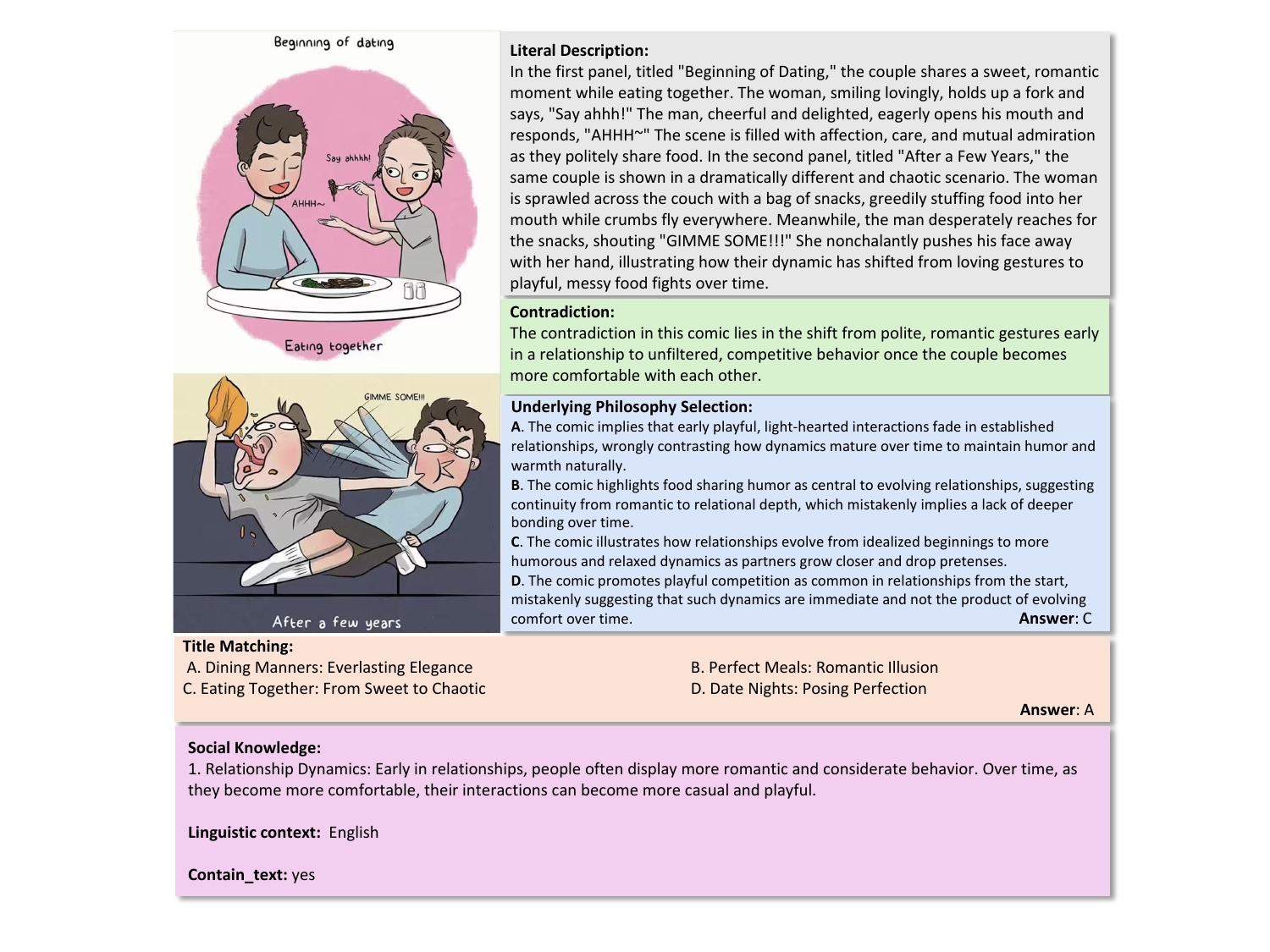}
    \caption{Sample Comic with All Annotated Tasks.}
    \label{fig:all annotated tasks}
\end{figure*}

\subsection{Annotation Prompts}
\label{sec:appendix_exp_rompts}
To balance efficiency and accuracy, we employ an AI-human collaborative pipeline for annotation. The AI assists in generating initial components, while human annotators refine outputs to ensure quality. The specific prompts used for AI generation are detailed in Fig.~\ref{fig:annotation_prompts}. We present a sample comic with all tasks in Fig. \ref{fig:all annotated tasks}.

\subsection{Detailed Data Distribution}
\label{sec:appendix_data_distribution}
Table~\ref{tab:attribute_distribution} provides a detailed breakdown of the dataset distribution based on different attributes, including Presence of Text, Social Norms, and Humor Categories.
To cluster images under humor categories, a compact and informative descriptive sentence is generated for each image based on its oracle description. These sentences are then processed using GPT-4 to identify thematic patterns, grouping them into distinct humor-related categories. This process results in 15 well-defined classes, each representing a unique type of humor. The diverse range of everyday life scenarios captured in our dataset provides a strong foundation for evaluating humor understanding in various contexts.

\begin{table}[h]
\begin{center}
\caption{Dataset Statistics Overview}
\label{tab:attribute_distribution}
\resizebox{\linewidth}{!}{
\begin{tabular}{l|ll|c}
\toprule

\multicolumn{1}{c|}{Attributes} & \multicolumn{2}{c|}{Subcategory} & \# of image \\
\midrule
    & \multicolumn{2}{c|}{No Text}   & 530   \\
    & \multicolumn{1}{l}{} & Linguistic-Dependent       & 370   \\
    \multirow{-3}{*}{Presence of Text}  & \multicolumn{1}{l}{\multirow{-2}{*}{\begin{tabular}[c]{@{}l|@{}}With \\ Text\end{tabular}}} & Transferable Text & 362\\

\midrule
                                  & \multicolumn{2}{c|}{Yes}   & 1091  \\
\multirow{-2}{*}{Social Norms}    & \multicolumn{2}{c|}{No}    & 171   \\

\midrule
\multirow{15}{*}{Humor Categories}  & \multicolumn{2}{c|}{Work Jokes \& Complaints}   & 233  \\
    & \multicolumn{2}{c|}{Internet Culture \& Technology}   & 202  \\
    & \multicolumn{2}{c|}{Fashion Trends \& Jokes}   & 112  \\    
    & \multicolumn{2}{c|}{Travel \& Adventure}   & 106  \\
    & \multicolumn{2}{c|}{Finance \& Money Matters}   & 100  \\ 
    & \multicolumn{2}{c|}{Sports \& Fitness}   &  92 \\
    & \multicolumn{2}{c|}{Food \& Culinary Experiences}   &  77 \\
    & \multicolumn{2}{c|}{Daily Life \& Routine Humor}   &  66 \\
    & \multicolumn{2}{c|}{Relationships \& Social Life}   & 62  \\
    & \multicolumn{2}{c|}{Entertainment \& Pop Culture}   &  53 \\
    & \multicolumn{2}{c|}{Education \& Student Life}   & 46  \\
    & \multicolumn{2}{c|}{Pet Behavior \& Humor}   &  37 \\
    & \multicolumn{2}{c|}{Health \& Wellness}   & 36  \\
    & \multicolumn{2}{c|}{Shopping \& Consumer Behavior}   &24\\
    & \multicolumn{2}{c|}{Holiday \& Seasonal Events}   & 16  \\

\bottomrule
\end{tabular}
}

\end{center}
\end{table}

\section{Experiments Details}

\subsection{Model Details}
Our experiments include both cutting-edge proprietary and open-source VLMs and LLMs, enabling a comprehensive evaluation across diverse architectures. For commercial VLMs, we use GPT-4o (\textit{gpt-4o-2024-08-06}) and GPT-4-Vision-turbo (\textit{gpt-4-turbo-2024-04-09})~\footnote{\url{https://platform.openai.com/docs/models/}}. 

Among open-source VLMs, our selection includes LLaVA-Next, covering 7B, 13B, and 72B parameter sizes~\cite{liu2024llavanext}, as well as LLaVA-1.5 in 7B and 13B variants~\cite{liu2023improved}. We also incorporate CogVLM2~\cite{hong2024cogvlm2}, Qwen2-VL with 7B and 72B versions~\cite{Qwen-VL}, and LLaVA-OneVision, which is available in 7B and 72B configurations~\cite{li2024llava}.

For LLMs, we use the Llama 3 instruction variant in both 8B and 70B sizes~\cite{llama3modelcard}, GPT-4 (\textit{gpt-4-0613}), DeepSeek-R1-70B (\textit{DeepSeek-R1-Distill-Llama-70B}), and Qwen2.5, available in 7B and 72B versions.

\subsection{Implementation Details}
\label{sec:appendix_implementation}
All commercial models are accessed through their official API, while open-sourced models are implemented using Hugging Face Transformers~\footnote{\url{https://huggingface.co/docs/transformers/en/index}}. Inference for GPT-3, GPT-4, GPT-4o, and GPT-4-Vision-Turbo is performed with a temperature of 1.0, while other models follow their default parameter settings or use greedy decoding. Experiments are conducted on A100 (80GB) and A6000 GPUs. 

For multiple-choice question (MCQ) evaluation, the models are explicitly instructed to directly output an option in the prompt. Answers are parsed using hard rules, and if no valid option is detected, a random choice is assigned. 
For generation task evaluation, we apply rouge-score~\footnote{\url{https://pypi.org/project/rouge-score/}} to compute ROUGE score, and calculate the BERT score using the official implementation~\footnote{\url{https://github.com/Tiiiger/bert_score}}. For GPT based evaluations for literal description and contradiction, we use \textit{gpt-3.5-turbo-0125} version. The prompts we used are shown in Fig.~\ref{fig:prompt_gpt_eval}.

\begin{figure}[t]
\centering
\Large
\renewcommand{\arraystretch}{1.5}  
\resizebox{1\linewidth}{!}{  
\begin{tabular}{p{0.7\textwidth}}  
\toprule

\textbf{Prompts for Literal Description:} \\
\midrule
- Candidate literal description: {gen}

- Reference literal description: {ref} \\

Task: You need to determine how accurately the above candidate literal description matches the given reference literal description of a comic narrative.\\

Using a scale from 1 to 5, rate the accuracy with which the candidate description matches the reference description, with 1 being the least accurate and 5 being the most accurate.\\
Please directly output a score by strictly following this format: [[score]], for example: Rating: [[3]]. \\  

\midrule

\textbf{Prompts for Contradiction:} \\
\midrule
Background: You are an impartial judge. You will be given a literal description of a comic that presents the same situation from two opposing perspectives, highlighting contradictions. You will also be provided with a gold-standard illustration as reference that effectively demonstrates these narrative contradictions.\\

Your task is to evaluate the quality of a generated illustration and determine whether it accurately depicts the narrative contradictions in the comic. Then, assign a score on a scale of 1 to 5, where 1 is the lowest and 5 is the highest, based on its quality.\\

- The literal description of the comic: {description}

- The reference contradiction illustration: {ref}

- The generated contradiction illustration: {gen} \\

Please directly output a score by strictly following this format: [[score]], for example: Rating: [[3]]. \\

\bottomrule
\end{tabular}
}
\caption{Prompts for GPT-based Evaluations}
\vspace{2mm}
\label{fig:prompt_gpt_eval}
\end{figure}

\subsection{Human Evaluations}
\label{sec:appendix_human_eval}
For literal description writing and contradiction generation tasks, we randomly select 40 samples from each task for human evaluation. Model outputs are anonymized and shuffled before being presented to the reviewer. Following~\cite{hwang-shwartz-2023-memecap}, the evaluation considers three key aspects:

\begin{itemize}[]
    \item {\textbf{Correctness}:} The model's output accurately conveys the narrative of the comic.
    \item {\textbf{Completeness}:} The model's output covers all important elements of the comic's narrative.
    \item {\textbf{Faithfulness}:} All content in the model's output is supported by the comic image, with no hallucinated information.
\end{itemize}

For literal description writing, we evaluate all three aspects, while for contradiction generation, only correctness and faithfulness are assessed.


\subsection{Model Finetuning Details for Deep Reasoning Tasks} \label{sec:appendix:finetune}

Our approach employs a weakly supervised textual data synthesis pipeline using powerful LLMs, such as GPT-4o, as a data generator. Instead of relying on paired image-text data, we replace comic images with textual descriptions of narratives, accompanied by corresponding reasoning questions. This approach leverages the advanced text generation capabilities of LLMs while eliminating the need for visual data during training.

\begin{figure}[t]
    \centering
    \def\arraystretch{1.5}
    \fontsize{9}{10}\selectfont
    \setlength{\tabcolsep}{1mm}
        \resizebox{1\linewidth}{!}{
    \begin{tabular}{p{0.45\textwidth}}
        \toprule
        \textbf{Data Generation Prompt:} \\
        \midrule
        You are a creative comic writer and a question-generation expert. Your task is to craft engaging narratives for ten two-panel comics. Each comic should depict the same situation from two contrasting perspectives, emphasizing contradictions or opposing viewpoints. The narratives should be thought-provoking, emotionally engaging, and capable of sparking discussions about some popular topics and events of daily life.

After crafting the comic narrative, generate two challenging multiple-choice questions:
- Moral question: Explore the underlying moral or philosophical lesson presented in the comic.
- Title question: Encourage thoughtful selection of a title that best captures the essence of the comic.

The questions should inspire deep reflection and provide meaningful answer choices that encourage nuanced thinking.

Below are several examples:\\
Example\_1 \\
Example\_2\\
...\\
Example\_10

Now generate five comic and questions. The output should strictly be a jsonlist, with output presented as a JSON object: 
\\
        \bottomrule
    \end{tabular}
    }
     \caption{Prompts used for Data Generation.}
    \label{fig:data generation}
    \vspace{2mm}
\end{figure}

\textbf{Data Generation.}
To construct the dataset, we manually select 10 diverse comic samples and utilize the few-shot learning ability of GPT-4o to generate 20,000 scene descriptions using the prompt shown in Fig.~\ref{fig:data generation}. A temperature of 1.0 is set for GPT-4o to encourage diversity in scene generation.

Next, based on these initial exemplars, GPT-4o synthesizes 20,000 contradictory comic scene descriptions. For each generated scene, additional prompts are used to create reasoning questions targeting symbolism selection and title matching tasks.


\begin{table*}[t!]
\centering
\begin{minipage}{\textwidth}
    \centering
    \Large
    \caption{Comparison of Training Hyperparameters for Qwen2-VL-7B, LLaVA-13B, and LLaVA-7B.
    }
    \label{tab:model_training_hyperparameters}
    \resizebox{\linewidth}{!}{%
    \begin{tabular}{l|c|c|c}
    \toprule
    \textbf{Hyperparameter} & \textbf{Qwen2-VL-7B} & \textbf{LLaVA-Next-13B} & \textbf{LLaVA-Next-7B} \\
    \midrule
    
    Per-device Train Batch Size & 4 & 4 & 4 \\
    Gradient Accumulation Steps & 4 & 4 & 4 \\
    
    Floating Point Precision & \texttt{bf16} & \texttt{bf16} & \texttt{bf16} \\
    Logging Steps & 25 & 25 & 25 \\
    Evaluation Strategy & Steps & Steps & Steps \\
    Evaluation Steps & 500 & 500 & 500 \\
    Save Steps & 500 & 500 & 500 \\
    LoRA Rank ($r$) & 256 & 128 & 32 \\
    LoRA Alpha & 1536 & 512 & 64 \\
    Number of Training Epochs & 5 & 5 & 5 \\
    LoRA Target Modules & \begin{tabular}[c]{@{}c@{}} \texttt{q\_proj, k\_proj, v\_proj, o\_proj,} \\ \texttt{gate\_proj, up\_proj, down\_proj} \end{tabular} 
                          & \begin{tabular}[c]{@{}c@{}} \texttt{q\_proj, k\_proj, v\_proj, o\_proj,} \\ \texttt{gate\_proj, up\_proj, down\_proj} \end{tabular} 
                          & \begin{tabular}[c]{@{}c@{}} \texttt{q\_proj, k\_proj, v\_proj, o\_proj,} \\ \texttt{gate\_proj, up\_proj, down\_proj} \end{tabular} \\
    \bottomrule
    \end{tabular}%
    }
\end{minipage}

\vspace{1cm}

\input{prompts_exp}
\end{table*}

\textbf{Finetuning Process.}
The generated dataset is then used to finetune \textit{the language components} of VLMs while keeping their visual perception modules unchanged. This targeted finetuning enhances the models’ comic understanding and reasoning abilities while preserving their original visual processing architecture. Table~\ref{tab:model_training_hyperparameters} presents the key training hyperparameters used for fine-tuning Qwen2, LLaVA-Next-7B, and LLaVA-Next-13B via the LoRA method.

\subsection{Evaluation Prompts} \label{sec:appendix_eval_prompts}
To ensure a fair evaluation, we design three distinct prompts, each independently crafted by different individuals to minimize biases introduced by prompt variations. Model performance is assessed using these prompts, and results are averaged across all tasks.

The prompts are designed to assess diverse aspects of model reasoning and understanding, covering Literal Description, Contradiction Generation, Symbolism Selection, and Title Matching are provided in Table~\ref{fig:prompt_exp_combined}.

\section{More Results}

\subsection{More Sample Outputs}
\label{sec:sample_outputs_appendix}

\begin{figure*}[b]
    \centering
    \includegraphics[width=1\linewidth, trim=30 10 30 80, clip]{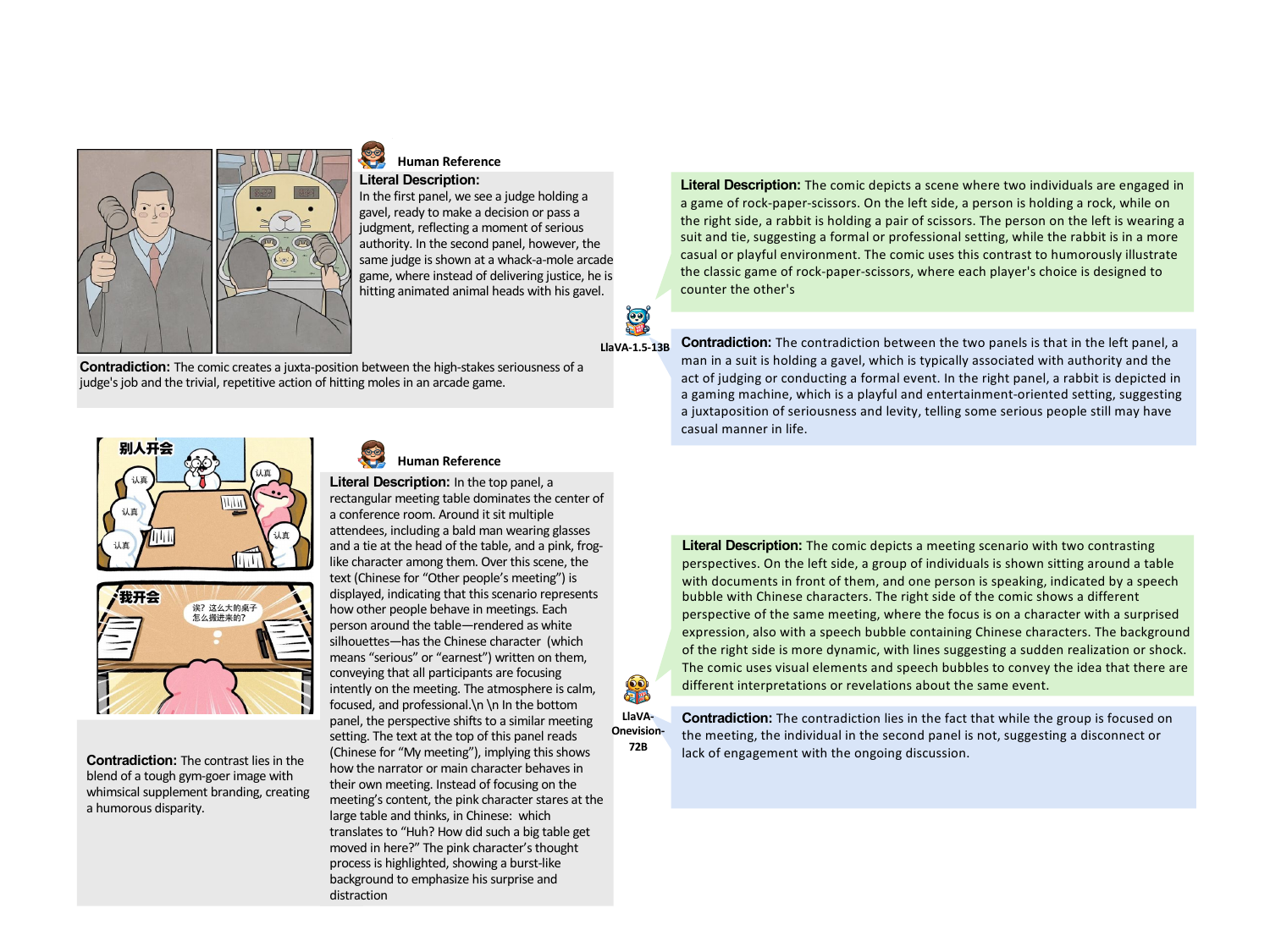}
    \caption{Sample outputs of model generated literal description and contradiction.}
    \label{fig:More samples}
\end{figure*}

\begin{figure*}
    \centering
    \includegraphics[width=1\linewidth, trim=30 70 30 10, clip]{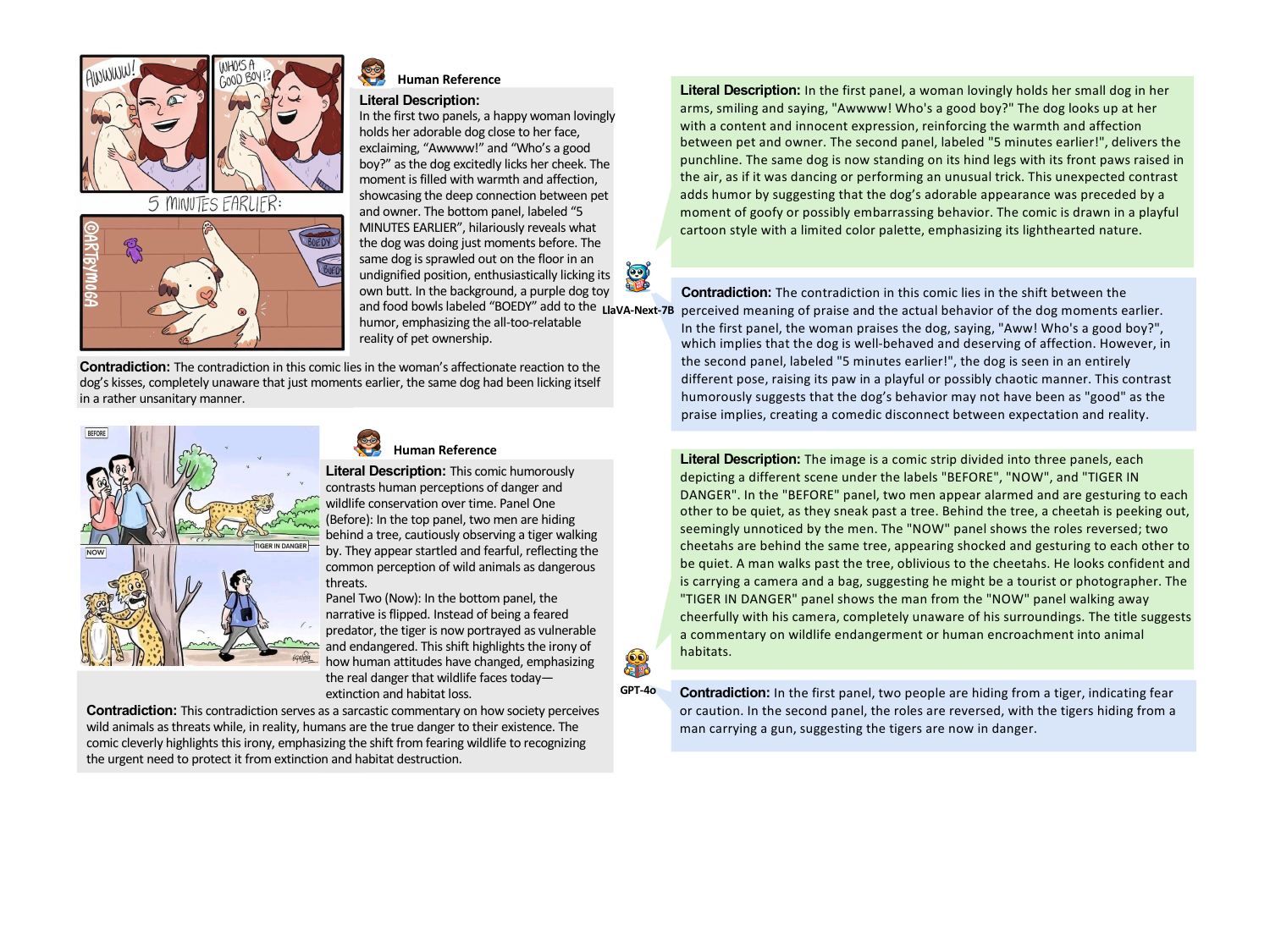}
    \caption{Sample outputs of model generated literal description and contradiction.}
    \label{fig:More samples2}
\end{figure*}

We present more randomly picked sample outputs on literal description and contradiction generation in Fig.~\ref{fig:More samples} and Fig.~\ref{fig:More samples2}.

\begin{table}[t!]
\centering
\caption{Comparison of model performance on deep reasoning tasks across different language contexts. Accuracy results for Symbolism Selection and Title Matching are reported for Language-independent, English-context, Chinese-context, and all images.}
\resizebox{\columnwidth}{!}
{
\begin{tabular}{l l c c}
\hline
\textbf{Model} & \textbf{Setting} & \textbf{Symbolism Acc.} & \textbf{Title Acc.} \\
\hline
\multirow{3}{*}{GPT-4o} 
  & Language-Free       & 81.12 & 81.65 \\
  & Chinese-Context  & 75.23 & 75.23 \\
  & English-Context  & 83.81 & 81.90 \\
  & All  & 80.38 & 80.62 \\
\hline
\multirow{3}{*}{LLaVA-OneVision-0.5B} 
  & Language-Free       & 40.62 & 42.21 \\
  & Chinese-Context  & 24.30 & 20.56 \\
  & English-Context  & 27.62 & 41.90 \\
  & All  & 36.87 & 38.53 \\
\hline
\multirow{3}{*}{LLaVA-OneVision-7B} 
  & Language-Free       & 69.67 & 72.96 \\
  & Chinese-Context  & 55.61 & 57.94 \\
  & English-Context  & 73.33 & 77.14 \\
  & All  & 67.64 & 70.81 \\
\hline
\multirow{3}{*}{LLaVA-OneVision-72B} 
  & Language-Free       & 81.23 & 79.92 \\
  & Chinese-Context  & 72.74 & 70.56 \\
  & English-Context  & 86.98 & 81.27 \\
  & All  & 80.30 & 78.48 \\
\hline
\multirow{3}{*}{Qwen2-VL-7B} 
  & Language-Free       & 73.59 & 73.81 \\
  & Chinese-Context  & 75.23 & 78.04 \\
  & English-Context  & 80.95 & 76.19 \\
  & All  & 74.48 & 74.72 \\
\hline
\multirow{3}{*}{Qwen2-VL-72B} 
  & Language-Free       & 78.79 & 80.81 \\
  & Chinese-Context  & 83.18 & 83.18 \\
  & English-Context  & 83.81 & 80.95 \\
  & All  & 79.98 & 81.25 \\
\hline
\end{tabular}
}
\label{tab:comparison}
\end{table}

\subsection{Impact of Language Context on Deep Reasoning Tasks}
\label{sec:appendix:language_influence}
We analyze how different language backgrounds influence a model’s ability to understand juxtaposition humor in deep reasoning tasks. Based on our statistical results in Table~\ref{tab:comparison}, we compare model performance across three settings: language-independent, Chinese-context, and English-context.

In the English-context setting, LLaVA-OneVision-72B and GPT-4o achieved the highest accuracy in deep reasoning tasks. Specifically, LLaVA-OneVision-72B attained 86.98\% accuracy in symbolism selection, while GPT-4o achieved 81.90\% in title matching.

In contrast, models performed differently in the Chinese-context setting. Qwen2-VL-72B consistently achieved the highest accuracy in both deep reasoning tasks. This result reinforces the superior understanding and reasoning ability of the Qwen2 series in Chinese, further validating our previous findings on text-included comics and Qwen's specialized performance in this context.

The language-independent setting yielded the most stable performance across all models. Since this setting primarily involves common scene humor, it does not require comprehension within a specific language. In this scenario, LLaVA-OneVision-72B achieved 81.23\% in the symbolism selection task, while GPT-4o attained 81.56\% in title matching.


%% file: prompts_exp.tex
\begin{minipage}{\textwidth}
    \centering
    \Large
    \caption{Example evaluation prompts used for Literal Description, Contradiction Generation, Symbolism Selection, and Title Matching. Three distinct prompts are designed for each task to minimize bias and ensure robust model evaluation.}
    \label{fig:prompt_exp_combined}
    \renewcommand{\arraystretch}{1.5}
    \captionsetup{skip=2mm}
    \resizebox{0.9\linewidth}{!}{
    \begin{tabular}{c | p{1.1\textwidth}}
    \toprule
    \textbf{Task} & \textbf{Prompts} \\
    \midrule
    \multirow{3}{*}{\textbf{Literal \newline Description}} 
        & \textbf{Prompt 1:} The given comic shows the same situation from two opposite sides with contradictions. Write a one-paragraph literal description to describe the narrative of the comic. \\ 
        & \textbf{Prompt 2:} Please literally describe the context of the image in detail. \\ 
        & \textbf{Prompt 3:} Give me a detailed literal description of the image. \\ 
    \midrule
    \multirow{3}{*}{\textbf{Contradiction Generation}} 
        & \textbf{Prompt 1:} The given comic shows the same situation from two opposite sides with contradictions. Write a short explanation to illustrate the contradiction of the two sides. \\ 
        & \textbf{Prompt 2:} Analyze the provided image, which is divided into two or more panels, each illustrating contrasting views of the same scenario. Describe the elements visible in each panel. Then concisely interpret how these elements convey contrasting perspectives in one or two sentences. Focus and only output the contradiction. \\ 
        & \textbf{Prompt 3:} Given an image with two or more panels showing a contrast relationship, describe the elements visible in each panel and concisely interpret the contradiction in one or two sentences. \\ 
    \midrule
    \multirow{3}{*}{\textbf{Underlying Symbolism Selection}} 
        & \textbf{Prompt 1:} The given comic shows the same situation from two opposite sides with contradictions. Which of the following options best represents the underlying Symbolism of the comic? \{MCQ Options\} Just output the choice. \\ 
        & \textbf{Prompt 2:} You are presented with an image divided into panels, each illustrating contrasting views of the same scenario. Which of the following options best represents the Symbolism of the image provided? \{MCQ Options\} Select the correct option by typing the corresponding letter (A, B, C, or D). \\ 
        & \textbf{Prompt 3:} Given an image with two or more panels showing contrast, select the best option representing the deep semantic of the image. \{MCQ Options\} Just output the correct option as (A, B, C, or D), no more explanation. \\ 
    \midrule
    \multirow{3}{*}{\textbf{Title  Matching}} 
        & \textbf{Prompt 1:} The given comic presents the same situation from two opposing perspectives, highlighting contradictions. Which of the following titles is most suitable for the comic? \{MCQ Options\} Output only the selected choice. \\ 
        & \textbf{Prompt 2:} You are presented with an image divided into two or more panels, each depicting contrasting perspectives of the same scenario. Which of the following title options best represents the given image? \{MCQ Options\} Select the correct option by typing the corresponding letter (A, B, C, or D). \\ 
        & \textbf{Prompt 3:} Given an image divided into two or more panels, a contrast relationship exists between the panels. Identify the best title from the following options that represents the image. \{MCQ Options\} Output only the corresponding letter (A, B, C, or D) without any additional explanation. \\ 
    \bottomrule
    \end{tabular}
    }

    \vspace{2mm}
\end{minipage}